\documentclass[9pt,twocolumn,twoside]{osajnl}
\definecolor{color1}{rgb}{0,0,0}
\definecolor{color2}{rgb}{0,0,0}
\definecolor{color3}{rgb}{0,0,0}
\providecommand{\journalref}{}
\providecommand{\OCIScodes}[1]{}
\providecommand{\ociscodes}[1]{}
\providecommand{\doi}[1]{}
\providecommand{\dates}[1]{}
\providecommand{\setprntext}[1]{}
\providecommand{\journallongtype}{}
\providecommand{\journalshorttype}{}

\providecommand{\journalname}{}

\providecommand{\subtitle}[1]{}

\usepackage{acro}
\usepackage{todonotes}
\usepackage{subcaption}
\usepackage{multirow} 
\usepackage{microtype}
\DeclareAcronym{ML}{
 short = ML,
 long  = Machine Learning
}

\DeclareAcronym{SOP}{
 short = SOP,
 long  = State of Polarization
}

\DeclareAcronym{VAE}{
 short = VAE,
 long  = Variational Autoencoder
}

\DeclareAcronym{DA}{
 short = DA,
 long  = Domain Adaptation
}

\DeclareAcronym{DNN}{
 short = DNN,
 long  = Deep Neural Network
}

\DeclareAcronym{tsne}{
 short = t-SNE,
 long  = t-Distributed Stochastic Neighbor Embedding
}

\DeclareAcronym{HPO}{
 short = HPO,
 long  = Hyperparameter Optimization
}

\DeclareAcronym{HP}{
 short = HP,
 long  = Hyperparameter
}

\DeclareAcronym{DL}{
 short = DL,
 long  = Deep Learning
}

\DeclareAcronym{OCSVM}{
 short = OCSVM,
 long  = One-Class Support Vector Machine
}

\DeclareAcronym{DBSCAN}{
 short = DBSCAN,
 long  = Density-Based Spatial Clustering of Applications with Noise
}

\DeclareAcronym{CNN}{
 short = CNN,
 long  = Convolutional Neural Network
}

\DeclareAcronym{SSL}{
 short = SSL,
 long  = Semi-supervised Learning
}

\DeclareAcronym{USL}{
 short = USL,
 long  = Unsupervised Learning
}

\DeclareAcronym{DSP}{
 short = DSP,
 long  = Digital Signal Processing
}

\DeclareAcronym{QoT}{
 short = QoT,
 long  = Quality of Transmission
}

\DeclareAcronym{SM}{
 short = SM,
 long  = Single-Mode
}

\DeclareAcronym{MM}{
 short = MM,
 long  = Multi-Mode
}

\DeclareAcronym{ROADM}{
 short = ROADM,
 long  = Reconfigurable Optical Add-Drop Multiplexer
}

\DeclareAcronym{CW-DFB}{
 short = CW-DFB,
 long  = Continuous Wave Distributed Feedback
}

\DeclareAcronym{ECL}{
 short = ECL,
 long  = External Cavity Laser
}

\DeclareAcronym{NPSV}{
 short = NPSV,
 long  = Numerical Polarization State Variation
}

\DeclareAcronym{FFT}{
 short = FFT,
 long  = Fast Fourier Transform
}

\DeclareAcronym{D}{
 short = D,
 long  = Dimensional
}

\DeclareAcronym{FC}{
 short = FC,
 long  = Fully Connected
}

\DeclareAcronym{RB}{
 short = RB,
 long  = Residual Block
}

\DeclareAcronym{BN}{
 short = BN,
 long  = Batch Normalization
}

\DeclareAcronym{DP}{
 short = DP,
 long  = Dropout
}

\DeclareAcronym{KL}{
 short = KL,
 long  = Kullback--Leibler
}

\DeclareAcronym{MSE}{
 short = MSE,
 long  = Mean Squared Error
}

\DeclareAcronym{CE}{
 short = CE,
 long  = Cross Entropy
}

\DeclareAcronym{LS}{
 short = LS,
 long  = Label Smoothing
}

\DeclareAcronym{FL}{
 short = FL,
 long  = Focal Loss
}

\DeclareAcronym{LR}{
 short = LR,
 long  = Learning Rate
}

\DeclareAcronym{TPE}{
 short = TPE,
 long  = Tree-structured Parzen Estimator
}
\DeclareAcronym{XGB}{
short= XGBoost,
long= eXtreme Gradient Boosting
}
\DeclareAcronym{AE}{
short=AE,
long=Autoencoder
}

\DeclareAcronym{AdamW}{
  short = AdamW,
  long  = Adaptive Moment Estimation with Decoupled Weight Decay
}

\DeclareAcronym{SGD}{
  short = SGD,
  long  = Stochastic Gradient Descent
}

\DeclareAcronym{RMSprop}{
  short = RMSprop,
  long  = Root Mean Square Propagation
}
\DeclareAcronym{SL}{
   short = SL ,
   long= Supervised Learning
}
\DeclareAcronym{WDM}{
short=WDM,
long= Wavelength Division Multiplexing
}
\DeclareAcronym{TL}{
 short= TL,
 long= Transfer Learning
}
\DeclareAcronym{M}{
short= M,
long= Million
} 
\setboolean{shortarticle}{false}

\title{
Variational Autoencoder Domain Adaptation for Cross-System Generalization in ML-Based SOP Monitoring 
}

\author[1,2,*]{Leyla Sadighi}
\author[3]{Stefan Karlsson}
\author[2]{Carlos Natalino}
\author[4]{Mojtaba Eshghie}
\author[2]{Fehmida Usmani}
\author[5]{Eoin Kenny}
\author[2]{Lena Wosinska}
\author[2]{Paolo Monti}
\author[2]{Marija Furdek}
\author[1]{Marco Ruffini}

\affil[1]{School of Computer Science and Statistics, IRIS research group, ADAPT Centre, Trinity College Dublin (TCD), Dublin, Ireland}
\affil[2]{Department of Electrical Engineering, Chalmers University of Technology, Gothenburg, Sweden}
\affil[3]{Micropol Fiberoptics AB, Stockholm, Sweden}
\affil[4]{Department of Computing Science, Umeå University, Umeå, Sweden}
\affil[5]{Asiera, Dublin, Ireland}
\affil[*]{sadighil@tcd.ie}

\begin{abstract}
Machine learning (ML) models trained to detect physical-layer threats on one optical fiber system often fail catastrophically when applied to a different system, due to variations in operating wavelength, fiber properties, and network architecture. To overcome this, we propose a Domain Adaptation (DA) framework based on a Variational Autoencoder (VAE) that learns a shared representation capturing event signatures common to both systems while suppressing system-specific differences. The shared encoder is first trained on the combined data from two distinct optical systems: a 21~km O-band dark-fiber testbed (System~1) and a 63.4~km C-band live metro ring (System~2). The encoder is then frozen, and a classifier is trained using labels from an individual system. The proposed approach achieves 95.3\% and 73.5\% cross-system accuracy when moving from System~1 to System~2 and vice versa, respectively. This corresponds to gains of 83.4\% and 51\% over a fully supervised Deep Neural Network (DNN) baseline trained on a single system, while preserving intra-system performance. 
\end{abstract}

\setboolean{displaycopyright}{false} 

\begin{document}

\maketitle

\section{Introduction}
\label{sec:introduction}
Optical fiber networks are the backbone of modern telecommunications infrastructure, carrying the vast majority of global data traffic. 
To keep pace with ever-increasing traffic demands, these networks are evolving rapidly: transmission is being extended into new spectral bands beyond the conventional C-band, 
and network operation is transitioning towards open, disaggregated and autonomous networks that combine various types of equipment from different vendors.
Modern optical networks are, therefore,  highly heterogeneous, and feature 
diverse deployments characterized by varying link lengths, topologies, and environmental conditions.

Beyond their primary role in communications, optical fibers are increasingly used for environment sensing and physical-layer threat monitoring applications~\cite{rafique_jlt_2018, allwood_2016_Sensors_Journal, pellegrini_jocn_2025}.
This enables a new role of optical fibers as ubiquitously available sensors of external disturbances, which helps identify potential threats, prevent service disruptions, and protect optical network integrity.

Optical fiber sensors enable high-precision environmental monitoring due to their sensitivity to various physical phenomena, including Rayleigh, Raman, and Brillouin scattering, as well as interferometric effects and polarization variations induced by external perturbations~\cite{lu_2019_APR}.
Among these sensing modalities, the \ac{SOP}is a particularly sensitive metric: external events such as vibrations or eavesdropping attempts alter the birefringence of the fiber, inducing characteristic, measurable variations in the \ac{SOP} of the propagating light.
The advantages of \ac{SOP}-based monitoring are bolstered by the widespread deployment of coherent optical transceivers in modern optical networks.
These devices inherently capture polarization-resolved optical field information through their dual-polarization receivers, enabling software-based \ac{SOP} estimation without requiring additional dedicated hardware~\cite{pellegrini_jocn_2025,carver_2024_nature}.
This capability makes \ac{SOP}-based network monitoring cost-effective and readily deployable in existing infrastructure.

Effectively interpreting \ac{SOP} signatures of different events remains a significant challenge.
Traditional monitoring approaches that rely on predetermined rules and fixed thresholds often fail to capture the complexity and variability of real-world fiber events, particularly those that introduce subtle, transient, or overlapping \ac{SOP} variations~\cite{rafique_jlt_2018}. 
Recent work has demonstrated that \ac{ML}-based classifiers applied to \ac{SOP} signatures can detect and categorize physical-layer events with high accuracy when trained and evaluated within the same optical system \cite{ls_ofc_2024}. 
However, the growing heterogeneity of optical networks, spanning different spectral bands, fiber types, link lengths, and environmental conditions, introduces significant variability in the physical-layer characteristics observed over different network segments, posing a fundamental challenge for data-driven monitoring approaches. This raises an important research question: \emph{can \ac{ML} models trained on one optical system maintain their accuracy when deployed on a system with fundamentally different physical characteristics (e.g. spectral bands, fiber links, and network topologies)?}

In our previous study~\cite{ls_ecoc_2025}, we demonstrated that an \ac{XGB} classifier trained on one system, which transmits a signal in the O-band over a 21\,km dark fiber (referred to as System~1), achieved only 9.08\% accuracy when tested on another system, transmitting in C-band over a 63.4\,km link in the Asiera metro network (referred to as System~2), which is worse than the theoretical random guessing.
As \ac{SOP} monitoring is characterized by systematic differences in spectral bands, fiber links, network topologies, and environmental conditions between systems, the distributions of the learned features diverge substantially, and the resulting domain shift causes unacceptable performance degradation~\cite{pan_tkde_2010}.

\ac{DA} techniques provide a framework for overcoming domain shift by learning representations that are invariant over source and target domains, that is, features that capture the essential characteristics of physical-layer events regardless of which optical system produced the data~\cite{pan_tkde_2010}. 
Several \ac{DA} strategies exist, each with distinct trade-offs. Direct \ac{TL} fine-tunes a source-pretrained model on the target domain, but requires labeled target data that may be scarce or costly in operational networks. Domain-adversarial methods encourage domain-invariant features by training against a domain discriminator, yet suffer from adversarial instability and risk discarding class-relevant information. Purely supervised approaches, regardless of architectural complexity, inherently overfit to source-domain statistics and offer no mechanism to handle unseen domain variation.
%
%
Among generative approaches, i.e., methods that model the underlying data distribution rather than directly mapping inputs to class labels, \acp{VAE} offer a compelling alternative~\cite{kingma_iclr_2014}. Their probabilistic, regularized latent space naturally separates domain-specific variation from shared structure, and when trained jointly on multiple systems, encourages geometrically similar representations for semantically similar events regardless of their system of origin. This makes \acp{VAE} particularly well suited to bridging domain shift between heterogeneous optical systems, yet their benefits for \ac{DA} in optical fiber monitoring have not been investigated.
%
%

In this paper, we extend our preliminary study~\cite{ls_ecoc_2025} by proposing a \ac{VAE}-based \ac{DA} framework for cross-system generalization in \ac{ML}-based \ac{SOP} signature classification. To the best of our knowledge, this is the first application of \ac{VAE}-based \ac{DA} to optical fiber monitoring.
In this work, we focus on classifying three physical-layer events from their \ac{SOP} signatures: the nominal undisturbed fiber state (\textit{relaxed}), unauthorized fiber tapping attempts (\textit{eavesdropping}), and gradual mechanical perturbations caused by fiber bending (\textit{soft bending}). We evaluate classification performance under four scenarios: two intra-system scenarios, $S_1$: System~1$\!\to\!$System~1 and $S_2$: System~2$\!\to\!$System~2, where the model is trained and tested on the same optical system, and two cross-system scenarios, $S_3$: System~1$\!\to\!$System~2 and $S_4$: System~2$\!\to\!$System~1, where the model is trained on one system and tested on a different one operating in a different spectral band, over a different fiber link, and within a different network topology. The cross-system scenarios are the primary focus of this work, as they directly probe whether \ac{ML} models can generalize across heterogeneous optical infrastructure.
%
%
%
To isolate the source of cross-system generalization gains, we establish a supervised \ac{DNN} baseline and a single-system \ac{VAE} ($\text{VAE}_{\text{sgl}}$) as reference points.
Both achieve strong intra-system accuracy yet collapse to below-chance under cross-system transfer, demonstrating that neither architectural sophistication nor probabilistic latent structure alone are sufficient to bridge the domain shift.
This motivates our proposed combined-system \ac{VAE} ($\text{VAE}_{\text{cmb}}$), which trains a shared encoder jointly on both systems through a two-phase design,
allowing the model to first learn the common patterns present in data from both systems before being taught to classify events. 
%
Our model achieves cross-system accuracy of 95.3\% for System~1$\!\to\!$System~2 and 73.5\% for System~2$\!\to\!$System~1, corresponding to gains of 83.4\% and 51\% over the \ac{DNN} baseline.
%
%
%

The remainder of the paper is organized as follows.
Section~\ref{sec:related_work} reviews the related work.
Section~\ref{sec:experiments_problem} presents the experimental testbed, problem formulation, and domain shift characterization. Section~\ref{sec:methodology} describes the proposed \ac{VAE}-based
\ac{DA} framework and the \ac{DNN} benchmark. Section~\ref{sec: ml} details the data preprocessing, \ac{HP} search spaces, and \ac{HPO} setup.
Section~\ref{sec: rslt} presents and discusses the results, while Section~\ref{sec: conclusion} concludes the paper.

\section{Related Work}
\label{sec:related_work}
The application of \ac{ML} to \ac{SOP}-based fiber monitoring has been proliferating over the recent years~\cite{pellegrini_jocn_2025, rode_oft_2025}.
Supervised \ac{SOP} signature classification has progressed from detection of harmful events on C-band laboratory data~\cite{ls_ofc_2024, ls_icton_2024} to robustness evaluation on real-world O-band deployments~\cite{ls_ecoc_2024}. 
Representative results from \ac{DL}-based approaches include accuracies of up to 98.57\% on a short-link (0.3~km round-trip) and 92.26\% on a long-link (21~km round-trip) metropolitan deployment~\cite{ls_JLT_2025}.
%
Other relevant \ac{SL} approaches include physical-layer threat monitoring via simple \ac{SOP} analyzers~\cite{tomasov_olt_2023}, vision-transformer-based anomaly classification from \ac{SOP} spectrograms~\cite{abdelli_jlt_2025}, abnormal activity detection on \ac{SM} and \ac{MM} fibers~\cite{karlsson_ofc_2023}, and \ac{CNN}-based fiber-bending eavesdropping detection~\cite{qin_ucom_2024}. 
Beyond \ac{SL} methods, \ac{SSL} and \ac{USL} anomaly detection using \ac{OCSVM} and \ac{DBSCAN} has been demonstrated~\cite{ls_tnsm_2025}. Cluster-based \ac{USL} eavesdropping detection has been proposed for \ac{WDM} systems~\cite{song_jocn_2024}, and \ac{DSP}-blind \ac{SOP}-based anomaly detection has been introduced for metropolitan fibers~\cite{minelli_ecoc_2023}.
%

Despite the high intra-system accuracy reported across these works, cross-system generalization remains a largely unresolved challenge.
The first systematic investigation of this problem for \ac{SOP}-based monitoring revealed that classifiers trained on one optical link experience catastrophic accuracy degradation when applied to a different link operating in a different spectral band~\cite{ls_ecoc_2025}.
This domain shift problem is well recognized in other optical networking problems solved through the application of \ac{ML}. For example, \ac{QoT} estimation models have been shown to fail to generalize across network topologies~\cite{rottondi_jocn_2021}, and analogous failures have been reported for cross-lightpath failure-cause identification~\cite{musumeci_jocn_2022}. Furthermore, transfer learning for \ac{QoT} prediction depends on source–target domain similarity~\cite{8386174}.
Several \ac{DA} and \ac{TL} strategies have been proposed to mitigate domain shift in optical networks, including model transfer for \ac{QoT} prediction~\cite{yu_jocn_2019}, domain adversarial adaptation with few labeled samples~\cite{cai_jocn_2024}, and \ac{TL} with image-encoded \ac{SOP} data for event classification under data scarcity~\cite{abdelli_jocn_2024}.
However, these approaches have primarily targeted \ac{QoT} estimation or general fiber fault management, while none has addressed the cross-system generalization problem for \ac{SOP}-based classification.

\acp{VAE}~\cite{kingma_iclr_2014} learn compact, structured representations of high-dimensional data by training a model to reconstruct its inputs through a constrained bottleneck. This bottleneck forces the model to capture only the most essential characteristics of the data, while the probabilistic regularization encourages similar inputs to be mapped to nearby points in the learned representation space. In the context of optical fiber monitoring, this means that \ac{SOP} signatures caused by the same physical event, such as eavesdropping, are represented similarly regardless of which optical system produced them, while system-specific differences, such as those arising from different spectral bands or fiber lengths, are naturally suppressed. 
%
Despite their demonstrated effectiveness in computer vision~\cite{ilse_midl_2020, louizos_iclr_2016} and signal processing~\cite{hsu_asru_2017}, \ac{VAE}-based \ac{DA} methods have not been applied to optical network monitoring. The core unresolved research challenge  in \ac{SOP}-based monitoring tackled in this work, i.e., learning representations that capture event-specific signatures while being invariant to the optical system that produced them, aligns directly with the strengths of \acp{VAE}: their regularized latent space naturally encourages inputs that share the same underlying cause to be mapped to similar representations, even when the raw measurements differ substantially due to system-specific factors. This paper leverages this property by introducing a \ac{VAE}-based \ac{DA} framework for cross-system \ac{SOP} signature classification, demonstrating that it can effectively bridge the severe domain shift between fundamentally different optical systems.
%
\section{Experimental Methodology and Problem Formulation}\label{sec:experiments_problem}
This section describes the two optical systems used for data collection, formalizes the cross-system generalization problem, and provides empirical evidence of the domain shift that motivates the proposed framework. 
\subsection{Optical Testbeds and \ac{SOP} Signature Collection}\label{subsec:testbed}

\begin{figure}[t]
    \centering
    \includegraphics[width=0.9\linewidth]{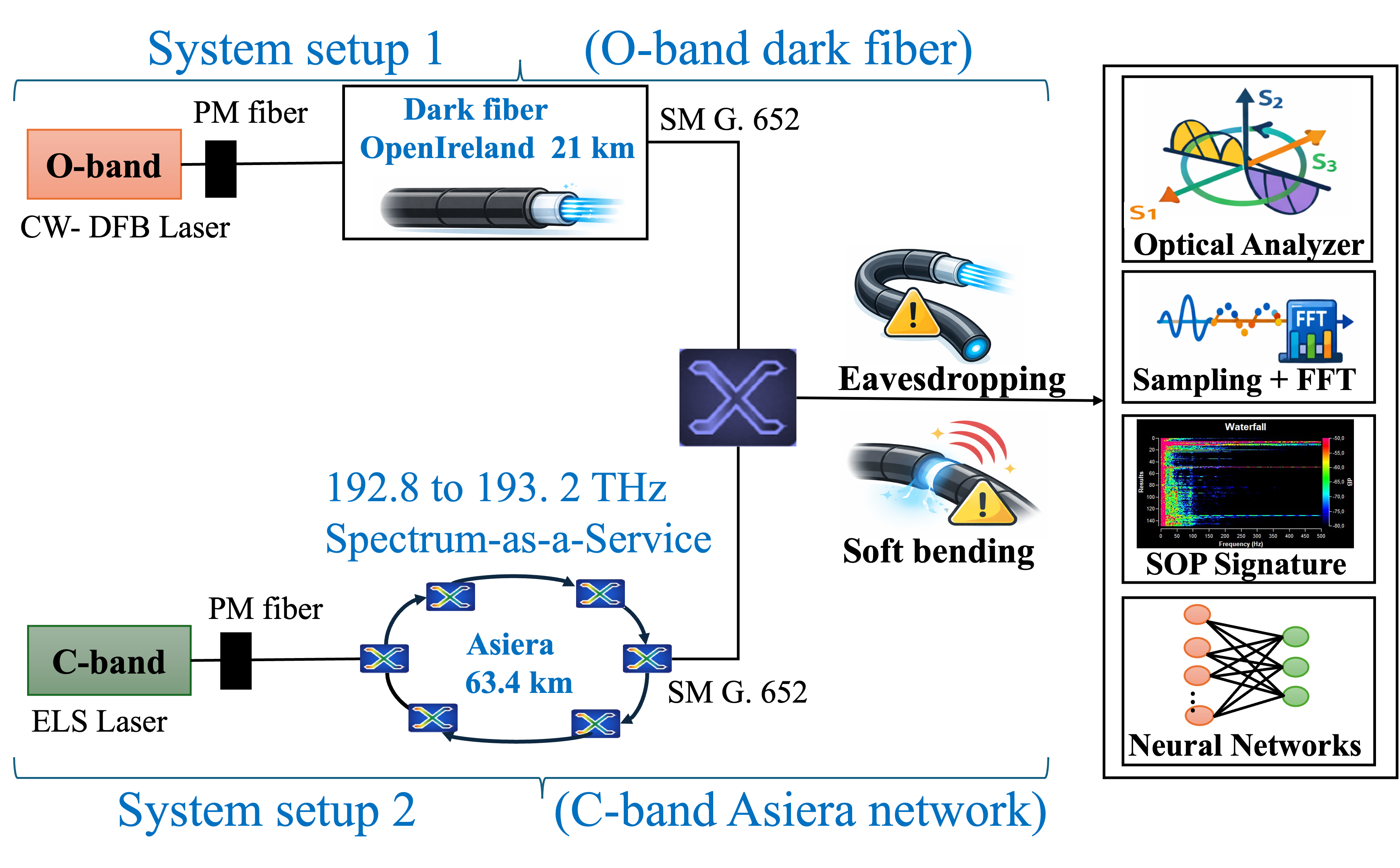}
    \caption{Schematic of experimental testbed comprising System~1 and System~2.
    }
    \label{fig:schm}
\end{figure}

The experimental setup used in this study, illustrated in Fig.~\ref{fig:schm}, comprises two independent fiber-optic measurement systems operating in distinct spectral bands.
The first, System~1, operates in the O-band over a 21~km round-trip dark fiber link accessed via the OpenIreland testbed \cite{OIR_2024}. It employs a \ac{CW-DFB} laser source and
standard \ac{SM} G.652 fiber. As a field-deployed fiber, this link introduces realistic impairments representative of production infrastructure.
The second, System~2, operates in the C-band over a live production metro ring network in
Dublin city, operated by Ireland's National Education and Research Network, Asiera (formerly HEAnet) \cite{HEAnet}. This network
spans 63.4~km of fiber across six \ac{ROADM} nodes and uses an \ac{ECL} source within a 400~GHz spectral window (192.8--193.2~THz), allocated as a Spectrum-as-a-Service slice. 
Both systems share a common optical analyzer, connected via a switch that selects one transmission link at a time. The selected link is subjected to controlled physical perturbations designed to replicate real-world tampering and 
eavesdropping attempts, each inducing distinct variations in the \ac{SOP} of the propagating signal.
%
%

We capture the \ac{SOP} traces at a 0.5~ms sampling interval over a 20-minute window for
each event, yielding 2.4~million samples per event and per spectral band.
From these traces, we derive a spectral feature
representation suitable for \ac{ML} analysis.
We begin by computing the \ac{NPSV}, a scalar metric that quantifies the spatial variation of the \ac{SOP} between successive sampling instants \cite{Karlsson_patent}. The measurement is based on two orthogonally polarized detection
channels, where the intensities of the horizontal ($I_1$) and vertical ($I_2$) polarization components are recorded independently. At each time instant $t$, we calculate $\Delta S(t)$ as $\Delta S(t) = I_1(t) - I_2(t)$, which reflects the phase-dependent interference between polarization modes and is sensitive to external perturbations on the fiber. The \ac{NPSV} at time slot~$t$ is defined as the spatial difference between consecutive samples:
$\text{NPSV}_t = \Delta S(t) - \Delta S(t-1)$.
A value of $\text{NPSV}_t \approx 0$ indicates a stable polarization state, while positive values reflect a shift of the polarization state toward horizontal dominance, and
negative values reflect a shift toward vertical dominance. The magnitude of $\text{NPSV}_t$ thus captures the speed and direction of \ac{SOP} transitions induced by physical
perturbations on the fiber.
The resulting \ac{NPSV} time series is partitioned into non-overlapping segments of 500~samples. A 512-point \ac{FFT} with a Hamming window \cite{podder_2014_hamming} is applied to each segment, producing a power spectrum of 512 frequency bins. Aggregated over the full observation window, this yields a two-dimensional spectral representation of 4{,}800 time segments $\times$ 512 frequency bins per event, constituting a unique \textit{SOP signature} that characterizes the
polarization response of the fiber under that specific physical perturbation.
%

This procedure is repeated independently for both spectral systems, producing two datasets that are used as inputs to the \ac{ML} pipeline. Each dataset comprises \ac{SOP} signatures from three physical-layer event classes, measured independently on both systems: relaxed (\textit{rlx}), representing the nominal undisturbed fiber state whose \ac{SOP} signature reflects only background environmental noise; eavesdropping (\textit{eav}), simulating unauthorized fiber tapping via optical couplers and producing characteristic \ac{SOP} perturbations 
from the induced bending stress; and soft bending (\textit{sbd}), representing gradual mechanical perturbations introduced by controlled fiber bending, which manifest as slower \ac{SOP} variations. The resulting System~1 and System~2 datasets comprise signatures $\{\textit{rlx}_1, \textit{eav}_1, \textit{sbd}_1\}$ and 
$\{\textit{rlx}_2, \textit{eav}_2, \textit{sbd}_2\}$, respectively, each containing 4{,}800 samples represented by 512-\ac{D} frequency feature vectors.

\subsection{Cross-System Generalization Problem Formulation}
\label{subsec:problem_statement}
Although both systems observe the same physical events and share the set of event classes $\mathcal{Y} = \{\textit{rlx},\, \textit{eav},\, \textit{sbd}\}$, their \ac{SOP} signatures are collected using distinct spectral bands, fiber lengths, and network topologies, inducing a \textit{domain shift} between the two systems. Formally, let $\mathcal{X}_S = \{(\mathbf{x}_i^S, y_i^S)\}_{i=1}^{n_S}$ and $\mathcal{X}_T = \{(\mathbf{x}_j^T, y_j^T)\}_{j=1}^{n_T}$ denote the annotated 
\ac{SOP} feature datasets from System~1 (source) and System~2 (target), with $n_S$ and $n_T$ denoting the number of labeled samples in each system, where $\mathbf{x} \in \mathbb{R}^{512}$ is a spectral feature vector and $y \in 
\mathcal{Y}$ its event class. The domain shift is characterized by:
\begin{align}
&P_S(\mathbf{x}) \neq P_T(\mathbf{x}),
\label{eq:marginal_shift} \\
&P_S(y \mid \mathbf{x}) \approx P_T(y \mid \mathbf{x}),
\label{eq:conditional_consistent}
\end{align}
where ~\eqref{eq:marginal_shift} states that the marginal input distributions differ due to system-specific polarization dynamics, and~\eqref{eq:conditional_consistent} reflects the physical expectation that identical events produce qualitatively consistent \ac{SOP} perturbations in both systems. This is the key assumption that makes generalization theoretically feasible.

\emph{The core problem is to learn a classifier $f:\mathbb{R}^{512} \rightarrow \mathcal{Y}$ that
transfers from $\mathcal{X}_S$ to $\mathcal{X}_T$ despite the marginal mismatch in
~\eqref{eq:marginal_shift}.}
The model must bridge this mismatch by learning a domain-invariant representation that inherently preserves the class consistency of ~\eqref{eq:conditional_consistent}.
%

%
%
\subsection{Empirical Characterization of Domain Shift}\label{subsec:DA}
\begin{figure}[t]
    \centering
    \includegraphics[width=0.95\linewidth]{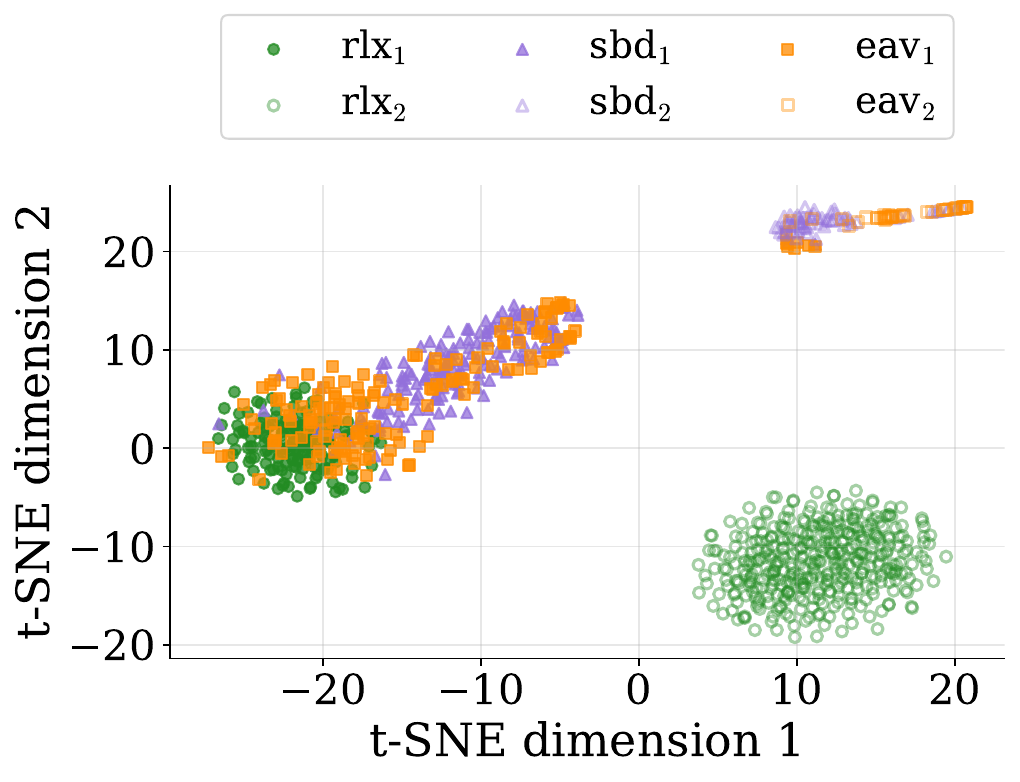}
    \caption{\ac{tsne} projection of 500 samples per system, colored by event class. Filled markers denote System~1 and open markers denote System~2. 
    }
    \label{fig:domain_shift}
\end{figure}

To confirm the presence of domain shift, we apply \ac{tsne}~\cite{van_2008_jml} to 500 training samples per system, projecting the 512-\ac{D} \ac{SOP} feature vectors onto a 2-\ac{D} embedding. Fig.~\ref{fig:domain_shift} colors each point by event class, using filled markers for System~1 and open markers for System~2.
The figure reveals two key observations. First, the filled and open markers occupy completely disjoint regions, confirming the marginal distribution mismatch of~\eqref{eq:marginal_shift}. Second, instances of the same physical event from different systems, e.g., \textit{rlx}$_1$ and \textit{rlx}$_2$, are spatially separated to a degree comparable to inter-class distances, directly explaining the near-chance cross-system accuracy of $9.08\%$ in scenario~$S_3$~\cite{ls_ecoc_2025}.

\section{PROPOSED METHODOLOGY}\label{sec:methodology}
The disjoint feature-space distributions revealed in Section~3.\ref{subsec:DA} indicate that direct cross-system transfer is highly challenging, motivating a representation learning 
approach that suppresses system-specific variation while preserving class discriminability. We first establish a supervised \ac{DNN} baseline to quantify the intra-system performance ceiling and the cross-system generalization gap, and then present the proposed \ac{VAE}-based framework designed to bridge this gap.
\subsection{\ac{FC} Residual \ac{DNN} Baseline}\label{subsec:dnn}
\begin{figure}[t]
    \centering
    \includegraphics[width=1\linewidth]{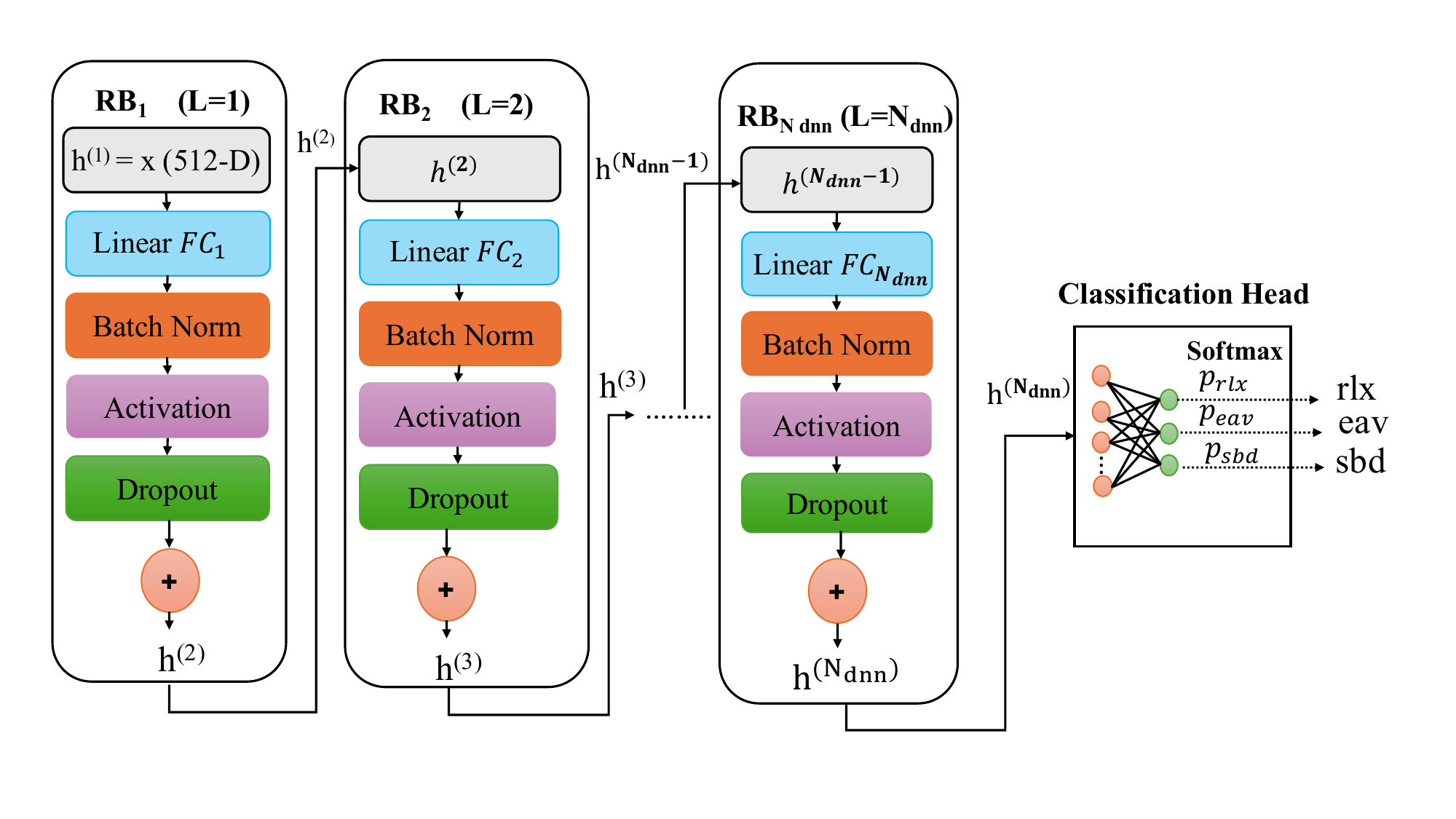}
    \caption{Architecture of the \acf{FC} \ac{DNN} baseline with sequential \acfp{RB}.}
    \label{fig:dnn model}
\end{figure}

As a supervised benchmark, we employ an \ac{FC} residual \ac{DNN} model that operates directly on the 512-\ac{D} \ac{SOP} feature vectors. The network architecture, as illustrated in Fig.~\ref{fig:dnn model}, follows a sequential \emph{encode-then-classify} structure: a backbone of $N_\mathrm{dnn}$ stacked \acp{RB} transforms the input representation, followed by a linear classification head that maps the final hidden state to class logits.
%
Each \ac{RB} at layer $\ell$ applies the transformation:
\begin{equation}\label{eq:resblock}
    \mathbf{h}^{(\ell+1)} = f\!\left(\mathrm{BN}\!\left(
    \mathbf{W}^{(\ell)}\mathbf{h}^{(\ell)} + \mathbf{b}^{(\ell)}\right)\right)
    + \mathbf{W}_{\mathrm{skip}}^{(\ell)}\,\mathbf{h}^{(\ell)}\text{,}
\end{equation}
\noindent where $\mathbf{h}^{(1)} = \mathbf{x}$ is the original 512-\ac{D} \ac{SOP} feature vector, and $\mathbf{h}^{(\ell)} \in \mathbb{R}^{d_\ell}$ for $\ell > 1$ is the hidden representation entering block $\ell$;
%
%
$\mathbf{W}^{(\ell)} \in \mathbb{R}^{d_{\ell+1} \times d_\ell}$ 
and $\mathbf{b}^{(\ell)} \in \mathbb{R}^{d_{\ell+1}}$ are the learnable weight
matrix and bias of the dense layer inside block $\ell$;
$\mathrm{BN}(\cdot)$ denotes \ac{BN}~\cite{BN_2015_ICML};
and $f(\cdot)$ is a non-linear activation function.
\noindent To prevent over-fitting to the relatively small per-system training sets, each block further applies \ac{DP}~\cite{DP_2014_JMLR} immediately after
the activation, stochastically zeroing a
fraction $p$ of the activations during training:
\begin{equation}\label{eq:dropout}
    \tilde{\mathbf{a}}^{(\ell)} = \mathrm{Drop}\!\left(
    f\!\left(\mathrm{BN}\!\left(\mathbf{W}^{(\ell)}\mathbf{h}^{(\ell)}
    + \mathbf{b}^{(\ell)}\right)\right),\, p\right)\text{,}
\end{equation}
\noindent where $p$ is treated as an \ac{HP} and should be selected by \ac{HPO}. The complete output of block $\ell$, incorporating both the regularized transformation and the skip connection ($\mathbf{W}_{\mathrm{skip}}$), is therefore:
\begin{equation}\label{eq:fullblock}
    \mathbf{h}^{(\ell+1)} = \tilde{\mathbf{a}}^{(\ell)}
    + \mathbf{W}_{\mathrm{skip}}^{(\ell)}\,\mathbf{h}^{(\ell)},
\end{equation}
\noindent 
The skip connection weight $\mathbf{W}_{\mathrm{skip}}^{(\ell)}$ takes one of
two forms depending on the layer dimensions:
\begin{equation}\label{eq:skip}
\mathbf{W}_{\mathrm{skip}}^{(\ell)} =
\begin{cases}
\mathbf{I} & \text{if } d_{\ell+1} = d_\ell
            \quad \text{(identity)},\\
\mathbf{W}_{p}^{(\ell)} \in \mathbb{R}^{d_{\ell+1} \times d_\ell}
  & \text{if } d_{\ell+1} \neq d_\ell
    \quad \text{(learned projection)}.
\end{cases}
\end{equation}
\noindent The outcome $\mathbf{h}^{(\ell+1)}$ is then passed directly as the input to the next block,
so that the representation is refined sequentially across all
$N_{\mathrm{dnn}}$ \acp{RB}.
After the final \ac{RB}, the output $\mathbf{h}^{(N_{\mathrm{dnn}}+1)}
\in \mathbb{R}^{d_n}$ constitutes a compact, task-relevant embedding of
the original \ac{SOP} feature vector, where $d_n$ denotes the width of
the last \ac{RB}. This embedding is passed to a linear classification
head, which projects it onto three output class scores:
\begin{equation}\label{eq:head}
    \mathbf{o} = \mathbf{W}_h\,\mathbf{h}^{(N_{\mathrm{dnn}}+1)} + \mathbf{b}_h,
    \qquad \mathbf{W}_h \in \mathbb{R}^{3 \times d_n},
\end{equation}
\noindent where $\mathbf{W}_h$ and $\mathbf{b}_h$ are learnable parameters and no activation function is applied. The class scores $\mathbf{o} \in \mathbb{R}^3$ are converted to class probabilities via the softmax function to obtain three output class scores corresponding to the event classes.
All model parameters of the residual backbone and the linear head are optimized jointly by minimizing a classification loss $\mathcal{L}_{\mathrm{dnn}}$ via gradient descent over the training set. The specific loss function (cross-entropy with optional label smoothing or focal loss) and the optimizer are treated as \acp{HP} and determined by \ac{HPO}, as detailed in Section~5.\ref{subsec: ml of dnn for hpo}.

%
\subsection{Proposed \ac{VAE} Framework for \ac{DA}}\label{subsec: vae architecture}
The \ac{DNN} baseline maps each input directly to a deterministic embedding optimized solely for classification on the source domain. As a consequence, the learned representation is shaped by system-specific characteristics and carries no mechanism to suppress domain-induced variation. 
To address this limitation, we propose a \ac{VAE}-based framework with a classification head that extends the deterministic encode-then-classify structure with a probabilistic encoder and a reconstruction decoder.
Unlike a standard \ac{AE}, which maps each input to a single-point representation in latent space $\mathbf{z}$, the \ac{VAE} encoder learns a conditional distribution $q_\phi(\mathbf{z}\mid\mathbf{x})$ over latent variables.
The \ac{KL} regularization term constrains this distribution toward a structured prior, encouraging a continuous and well-regularized latent space. Consequently, geometrically similar latent representations correspond to semantically similar inputs irrespective of their domain of origin, provided the encoder is exposed to both distributions during training.
The jointly trained classification head further ensures that the learned representation remains directly actionable for event detection, making the framework well-suited for cross-system transfer.

\begin{figure}[t]
    \centering
    \includegraphics[width=1\linewidth]{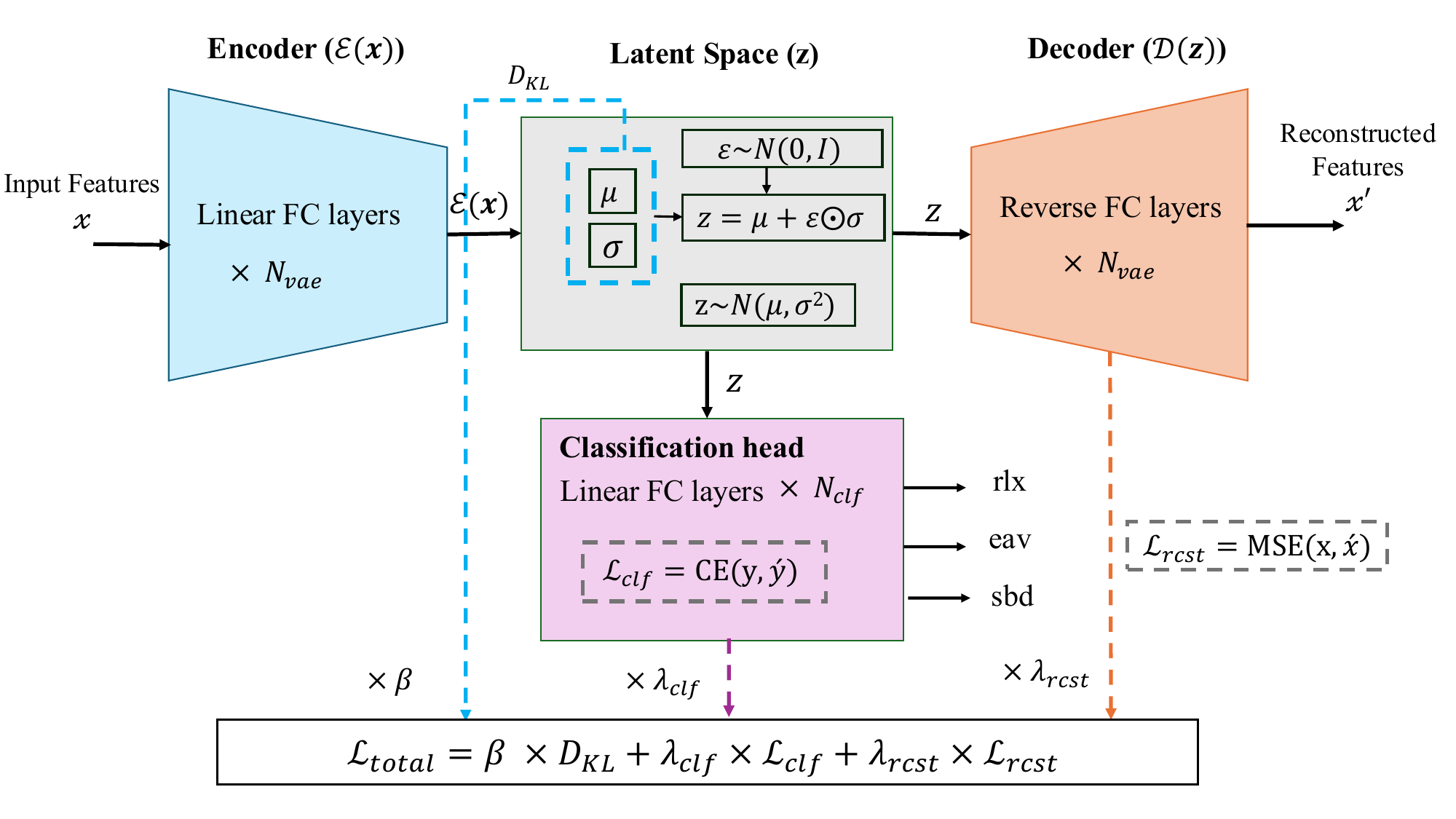}
    \caption{
    General architecture of the proposed \ac{VAE} framework
    comprises three components: an encoder, a decoder, and a classification head.
             }
    \label{fig:vae model}
\end{figure}

The proposed framework, illustrated in Fig.~\ref{fig:vae model}, comprises three components: a probabilistic encoder, a decoder, and a classification head. Given an input \ac{SOP} feature vector $\mathbf{x} \in \mathbb{R}^{512}$, the encoder $\mathbf{\mathcal{E}}(\mathbf{x})$, comprising $N_{vae}$ stacked \ac{FC} layers with \ac{BN} and \ac{DP} regularization, maps $\mathbf{x}$ to the parameters of a posterior Gaussian distribution over a low-\ac{D} latent space:
\begin{equation}\label{eq:vae_enc}
    \boldsymbol{\mu},\, \log\boldsymbol{\sigma}^2 = \mathcal{E}(\mathbf{x}),
    \qquad
    q_\phi(\mathbf{z} \mid \mathbf{x})
    = \mathcal{N}\!\left(\boldsymbol{\mu},\,\mathrm{diag}
      (\boldsymbol{\sigma}^2)\right),
\end{equation}
\noindent where $\boldsymbol{\mu},\boldsymbol{\sigma}^2 \in \mathbb{R}^{d_z}$ are the latent mean and variance vectors, and $d_z$ is the latent dimensionality selected by \ac{HPO}. A latent sample is drawn via the reparameterization trick,
\begin{equation}\label{eq:reparam}
    \mathbf{z} = \boldsymbol{\mu}
               + \boldsymbol{\varepsilon} \odot \boldsymbol{\sigma},
    \qquad \boldsymbol{\varepsilon} \sim \mathcal{N}(\mathbf{0}, \mathbf{I}),
\end{equation}
\noindent which preserves differentiability through the sampling operation and enables end-to-end training. The decoder $\mathcal{D}(\mathbf{z})$, a symmetric \ac{FC} network of $N_{vae}$ layers, reconstructs the input as $\mathbf{x}^{\prime} \in \mathbb{R}^{512}$, while the classification head maps $\mathbf{z}$ through $N_{\mathrm{clf}}$ \ac{FC} layers to class probabilities over $\mathcal{Y} = \{\textit{rlx},\,\textit{eav},\,\textit{sbd}\}$ via softmax.
The activation function and all layer dimensions are treated as \acp{HP} and will be determined by \ac{HPO}. 
All three loss components are optimized jointly through:
\begin{equation}\label{eq:vae_loss}
\begin{split}
    \mathcal{L}_{\mathrm{total}}
    = \;&\beta \cdot D_{\mathrm{KL}}\!\left(
        q_\phi(\mathbf{z}\mid\mathbf{x})
        \,\|\,\mathcal{N}(\mathbf{0},\mathbf{I})\right)\\
    +\;&\lambda_{\mathrm{rcst}}\cdot
      \mathcal{L}_{\mathrm{rcst}}(\mathbf{x},\mathbf{x}^{\prime})
    + \lambda_{\mathrm{clf}}\cdot\mathcal{L}_{\mathrm{clf}}(\hat{y},y),
\end{split}
\end{equation}
\noindent where $D_{\mathrm{KL}}$ measures the divergence between the approximate posterior $q_\phi(\mathbf{z}\mid\mathbf{x})$, inferred by the
encoder, and the standard Gaussian prior
$p(\mathbf{z}) = \mathcal{N}(\mathbf{0},\mathbf{I})$, regularizing the latent space toward a compact and well-structured distribution;
$\mathcal{L}_{\mathrm{rcst}} = \frac{1}{B}\sum_{i=1}^{B}
\|\mathbf{x}_i - \mathbf{x}_i^{\prime}\|^2$ is the \ac{MSE} over a mini-batch of $B$ samples between the input and its reconstruction, ensuring that $\mathbf{z}$ retains sufficient information to faithfully recover the original \ac{SOP} features; and $\mathcal{L}_{\mathrm{clf}} = -\frac{1}{B}\sum_{i=1}^{B} \log\hat{p}_{i,y_i}$ is the unweighted \ac{CE} loss over the same mini-batch, where $\hat{p}_{i,y_i}$ is the predicted probability assigned to the true class $y_i$ of sample $i$, enforcing class discriminability directly within the latent space. The weighting coefficient $\beta$ controls the strength of the \ac{KL} regularization relative to the other objectives,
$\lambda_{\mathrm{rcst}}$ balances reconstruction fidelity against the classification pressure, and $\lambda_{\mathrm{clf}}$ governs the contribution of the supervised signal in shaping the latent representation.
All three weighting coefficients, together with all remaining \acp{HP}, are determined through Bayesian \ac{HPO} using the Optuna framework, as described in Section~5.\ref{subsec: ml of vae for hpo}.
%

The framework is instantiated in two operational modes, each addressing a distinct generalization objective. Both modes share the same architectural template and loss formulation; their distinction lies solely in the scope of the input distribution the encoder is designed to handle.
\subsubsection{
Single-System \ac{VAE} ($\text{VAE}_\text{sgl}$): Per-System Reference Mode 
}\label{subsubsec: single vae}
In the $\text{VAE}_\text{sgl}$ mode, two independent \ac{VAE} models are constructed, one per system, each with its own encoder, decoder, and classification head. Each model produces a latent space conditioned exclusively on the statistics of its source domain, yielding strong intra-system discriminability under $S_1$ and $S_2$ and serving as a per-system performance reference. Since the two models are structurally isolated, their latent spaces are inherently domain-specific and carry no mechanism to bridge the marginal distribution mismatch of~\eqref{eq:marginal_shift}, making cross-system scenarios $S_3$ and $S_4$ a direct probe of how much class-discriminative structure is incidentally preserved under distribution shift.
\subsubsection{Combined-System \ac{VAE} ($\text{VAE}_\text{cmb}$): Combined Cross-System Training Mode}\label{subsubsec: combined vae}
In the $\text{VAE}_\text{cmb}$ mode, a single shared \ac{VAE} is constructed whose encoder simultaneously processes inputs from both systems, producing a unified latent space $\mathbf{z}$ that must account for the full cross-domain variability. The joint pressure from the reconstruction and \ac{KL} terms in~\eqref{eq:vae_loss} encourages the encoder to suppress system-specific variation and organize $\mathbf{z}$ around the class-discriminative structure that is consistent over both systems, in accordance with~\eqref{eq:conditional_consistent}. A separate classification head is trained and attached for each evaluation scenario, allowing the transferability of the shared latent representation to be probed independently under all four intra- and cross-system conditions.
\section{\ac{ML} Analysis}\label{sec: ml}
This section describes the data preprocessing pipeline, the \ac{HP} search spaces for the \ac{DNN} and \ac{VAE} frameworks, and the Bayesian \ac{HPO} setup used to identify optimal configurations.
\subsection{Data Preprocessing}\label{subsec: data preprocessing}
Since the quality of the raw data directly affects the reliability of the subsequent \ac{ML} analysis, the following preprocessing steps are applied on the collected data in Section~3.\ref{subsec:testbed}.
\subsubsection{Step 1: System~2 Activity Filtering}\label{subsub: filterind C band}
Since physical perturbations are applied manually and intermittently during data collection, the \textit{eav}$_2$ and \textit{sbd}$_2$ recordings naturally contain intervals in which the fiber remains relaxed between consecutive events. 
These segments are removed to retain only samples corresponding to active physical disturbances. This filtering step enhances the separability of each class's event-specific spectral characteristics from the baseline \textit{rlx}$_2$ state. 
After filtering, the System~2 dataset becomes imbalanced, and contains 644 \textit{eav}$_2$ samples and 773 \textit{sbd}$_2$ samples, 
while all 4{,}800 \textit{rlx}$_2$ samples are retained unchanged.
\subsubsection{Step 2: Domain-Aware Stratified Dataset Partitioning}\label{subsubsec: stratification}
Following preprocessing, the System~1 and System~2 datasets are merged 
into a unified dataset of 20{,}617 samples, each represented by a 
512-\ac{D} spectral feature vector with two associated labels: an event 
class label $y \in \mathcal{Y}=\{0,1,2\}$ and a domain label 
$d \in \{0,1\}$ identifying the source system. The dataset is then 
partitioned using two-stage stratified sampling on the joint key $(y,d)$, 
yielding 14{,}431 training (70\%), 3{,}093 validation (15\%), and 3{,}093 
test (15\%) samples, with identical class and domain proportions preserved 
across all splits, including the System~2 class imbalance. Per-system views 
are extracted from each split to support all four evaluation scenarios 
($S_1$--$S_4$) without re-splitting or data duplication. Since all models 
share the same splits, cross-model comparisons remain directly comparable. 
Table~\ref{tab:dataset_summary} summarizes the resulting sample counts.
\begin{table}[t]
\centering
\caption{Dataset and sample count summary per class and their allocation for
training, validation, and test splits.}
\label{tab:dataset_summary}
\small
\begin{tabular}{llccccc}
\toprule
\textbf{System} & \textbf{Class} & \textbf{Total} & \textbf{Train} & \textbf{Val} & \textbf{Test} \\
\midrule
\multirow{3}{*}{System~1 (O-band)} 
 & \textit{rlx}$_1$ & 4{,}800 & 3{,}360 & 720 & 720 \\
 & \textit{eav}$_1$ & 4{,}800 & 3{,}360 & 720 & 720 \\
 & \textit{sbd}$_1$ & 4{,}800 & 3{,}360 & 720 & 720 \\
\midrule
\multirow{3}{*}{System~2 (C-band)} 
 & \textit{rlx}$_2$ & 4{,}800 & 3{,}360 & 720 & 720 \\
 & \textit{eav}$_2$ &    644  &   450   &  97 &  97 \\
 & \textit{sbd}$_2$ &    773  &   541   & 116 & 116 \\
\midrule
\multirow{3}{*}{\textbf{Combined}} 
 & O-band & 14{,}400 & 10{,}080 & 2{,}160 & 2{,}160 \\
 & C-band &  6{,}217 &  4{,}351 &   933   &   933   \\
 & \textbf{All} & \textbf{20{,}617} & \textbf{14{,}431} & \textbf{3{,}093} & \textbf{3{,}093} \\
\bottomrule
\end{tabular}
\end{table}
\subsubsection{Step 3: Feature Normalization}\label{subsubsec: normalization}
Finally, z-score standardization is applied to each of the spectral features 
to ensure stable and unbiased model training. For a feature value $x$, the 
normalized value $\hat{x}$ is computed as $\hat{x} = \frac{x - \mu_{\text{train}}}{\sigma_{\text{train}}}$,
where $\mu_{\text{train}}$ and $\sigma_{\text{train}}$ are the feature-wise mean and standard deviation computed exclusively from the training set. The same 
statistics are then applied to normalize the validation and test sets.
\subsection{\ac{ML} Analysis of \ac{DNN} Model for \ac{HPO}}\label{subsec: ml of dnn for hpo}
The \ac{DNN} model exposes two categories of \acp{HP}: architectural and optimization. The architectural \acp{HP} govern the depth and width of the residual backbone. 
The number of \acp{RB}, $N_\mathrm{dnn}$, is varied over $\{2,3,4,5\}$. The width of the first hidden layer is searched over $\{256, 320, \ldots, 768\}$ (step 64).
%
The activation function $f(\cdot)$ in each \ac{RB} is selected from $\{\mathrm{GELU},\,\mathrm{ReLU},\,\mathrm{ELU},\,\mathrm{SiLU}\}$, the per-block \ac{DP} probability $p$ is searched over $[0.10,\,0.50]$, and \ac{BN} is applied with fixed default parameters (momentum $= 0.1$, $\varepsilon = 10^{-5}$, learnable affine transform).
%

The optimization \acp{HP} include the loss function,
optimizer, learning rate $\eta$, weight decay $\lambda$,
\ac{LR} schedule, and mini-batch size. The loss function is selected from two candidates: \ac{CE} with optional
\ac{LS} (smoothing parameter $\alpha \in [0.0,\,0.15]$,
where $\alpha=0$ recovers standard \ac{CE}), which
controls the degree of prediction confidence; and \ac{FL}
(focusing parameter $\gamma \in [0.5 \,5.0]$)~\cite{lin_2017_focal}, which down-weights easy examples to focus training on hard or
misclassified samples.
%
The optimizer is chosen from $\{\mathrm{AdamW},\,\mathrm{SGD\text{-}Nesterov},\,\mathrm{RMSprop}\}$:
\ac{AdamW} decouples weight decay from the gradient update and adapts per-parameter \acp{LR}, making it well-suited to sparse or heterogeneous gradients; \ac{SGD} with Nesterov momentum offers strong generalization when carefully tuned; and \ac{RMSprop} adapts step sizes via a running average of squared gradients, which can stabilize training under noisy or non-stationary objectives.
The \ac{LR}, i.e. $\eta$ is searched over $[10^{-4},\,5\times10^{-3}]$ (log-uniform) and weight decay $\lambda$ over $[10^{-6},\,10^{-3}]$ (log-uniform); momentum is additionally searched over $[0.80,\,0.99]$ when \ac{SGD} or \ac{RMSprop} is selected.
The \ac{LR} schedule is chosen between a linear warm-up followed by cosine annealing (warm-up duration $\in [5,20]$ epochs) and a plateau-based reduction (reduce on plateau, factor~$0.5$). Finally, the mini-batch size is selected from $\{32, 64, 128, 256\}$.

\subsection{\ac{ML} Analysis of \ac{VAE} Model for \ac{HPO}}
\label{subsec: ml of vae for hpo}
The \ac{VAE} search space is organized into three categories: architectural, loss-weighting, and optimization \acp{HP}.

The \textit{architectural} \acp{HP} cover the encoder depth (number of hidden layers), width (number of neurons in each layer), the latent
dimensionality, activation function, and regularization strategy. 
%
%
The number of hidden layers $N_{vae}$ is searched over $\{1, 2, 3, 4\}$, and each hidden layer width is sampled independently from $\{64, 128, 192, \ldots, 768\}$ (step~64, 12 values).
The latent dimensionality $d_z$ is selected from 
$\{8, 16, 32, 64, 128, 256\}$, the activation function is selected from 
$\{\mathrm{LeakyReLU},\,\mathrm{ReLU},\,\mathrm{GELU},\,
\mathrm{ELU},\,\mathrm{SiLU}\}$, \ac{BN} is treated as a binary \ac{HP} (enabled or disabled), and the \ac{DP} probability $p$ is searched over $[0.0,\,0.6]$ (uniform).

The \textit{loss-weighting} \acp{HP} are specific to the \ac{VAE} objective of~\eqref{eq:vae_loss}: the \ac{KL} weight $\beta$ is searched over $[10^{-4},\,5.0]$ (log-uniform), the reconstruction weight $\lambda_{\mathrm{rcst}}$ over $[0.01,\,5.0]$ (uniform), and the classification weight $\lambda_{\mathrm{clf}}$ over $[0.1,\,5.0]$ (uniform). The $\beta$ warm-up duration is searched over $[0,\,60]$ epochs, where a value of zero disables cosine annealing entirely and allows \ac{HPO} to determine whether gradual \ac{KL} introduction is beneficial for a given configuration.
%

The \textit{optimization} \acp{HP} govern the the optimizer,
learning rate $\eta$, weight decay $\lambda$, mini-batch
size, and \ac{LR} schedule. 
%
The optimizer is selected from $\{\mathrm{AdamW},\,\mathrm{Adam},\,\mathrm{SGD\text{-}Nesterov},\,\mathrm{RMSprop}\}$, with momentum additionally searched over $[0.80,\,0.99]$ when \ac{SGD} or \ac{RMSprop} is selected. The learning rate $\eta$ is searched over 
$[10^{-5},\,10^{-2}]$ (log-uniform), weight decay $\lambda$ over $[10^{-7},\,10^{-2}]$ (log-uniform), mini-batch size from $\{32, 64, 128, 256\}$. 
The \ac{LR} schedule is selected from three candidates: plateau-based reduction (factor~$0.5$, patience~$10$ epochs), cosine annealing, and one-cycle \ac{LR}.

\subsection{\ac{HPO} Setup and \ac{HP} Search Space}
\label{subsec: hpo space}
\acp{HP} for the \ac{DNN} and \ac{VAE} frameworks are tuned using \emph{Optuna} with a \ac{TPE} sampler~\cite{optuna_2019_ACM}, and a Median Pruner (10 startup trials, 15 warm-up steps) to terminate unpromising trials early.
For the \ac{DNN} and $\text{VAE}_{\text{sgl}}$, two independent optimization rounds are conducted per model: one optimized on the System~1 training and validation sets (used for $S_1$ and $S_3$) and one on the System~2 counterparts (used
for $S_2$ and $S_4$), to maximize source-domain
validation accuracy.
For $\text{VAE}_{\text{cmb}}$, a single optimization round is conducted on the combined system training and validation sets, to maximize the mean of the System~1 and System~2 validation accuracies, ensuring that the model generalizes well to both domains simultaneously rather than
a single one.
Each study runs for 100 trials, with each trial training for at most 150 epochs, subject to early stopping with patience of 20 epochs. Following \ac{HPO}, the best configuration identified for each study is used to
retrain a final model from scratch on the full training set, with a maximum of 500 epochs and early-stopping patience of 40 epochs.
All experiments were implemented in Python using the PyTorch \ac{DL} framework (v$\geq$2.0)~\cite{Pytorch_2019_NeurIPS} and executed on a
system equipped with NVIDIA H100 80GB GPUs.
\section{Results and Discussion}\label{sec: rslt}
This section presents and discusses the experimental results for four evaluation scenarios: 
intra-system $S_1$ and $S_2$, and cross-system $S_3$ and $S_4$. Results are reported in two stages: \ac{HPO}-selected architectures and training dynamics are analyzed first, followed by a comprehensive performance evaluation.
%
%
%
\subsection{Best Selected \acp{HP}}\label{subsec: hpo rslt}
%
%
\subsubsection{Best-Found \ac{DNN} Architectures}\label{subsubsec: dnn hpo rslt}
Table~\ref{tab:dnn_best} summarizes the \ac{HPO}-selected architecture and training configuration for the System~1 and System~2 of \ac{DNN} models. 

\begin{table}[t]
\centering
\caption{\ac{HPO}-selected configuration for the best-found \ac{DNN} models.}
\label{tab:dnn_best}
\small
\setlength{\tabcolsep}{4pt}
\begin{tabular}{lcc}
\toprule
\textbf{Hyperparameter} & \textbf{System~1} & \textbf{System~2} \\
                        & \textbf{($S_1$, $S_3$)} & \textbf{($S_2$, $S_4$)} \\
\midrule
\multicolumn{3}{l}{\textbf{Architecture}} \\
 $N_\mathrm{dnn}$  & 5 & 3 \\
Layer widths   & $512\!\to\!256\!\to\!64^{\times4}\!\to\!3$
               & $512\!\to\!704\!\to\!368\!\to\!92\!\to\!3$ \\
Parameters     & 161{,}475 & 657{,}147 \\
Activation $f(\cdot)$  & SiLU & SiLU \\
Dropout $p$            & 0.233 & 0.267 \\
\midrule
\multicolumn{3}{l}{\textbf{Optimization}} \\
Loss                         & \ac{CE}+\ac{LS} & \ac{CE}+\ac{LS} \\
\ac{LS} $\alpha$     & 0.085 & 0.082 \\
Optimiser                    & SGD (Nesterov) & AdamW \\
Learning rate $\eta$         & $3.36\!\times\!10^{-3}$ & $3.84\!\times\!10^{-4}$ \\
Weight decay $\lambda$       & $1.54\!\times\!10^{-6}$ & $3.99\!\times\!10^{-6}$ \\
Momentum                     & 0.803 & --- \\
\ac{LR} schedule             & Plateau ($\times$0.5) & Plateau ($\times$0.5) \\
Batch size                   & 32 & 128 \\
\midrule
\multicolumn{3}{l}{\textbf{Result}} \\
Best val.\ accuracy          & 85.79\% & 97.64\% \\
\bottomrule
\end{tabular}
\end{table}

The two models differ in both their depth and width, reflecting the distinct characteristics of each domain. The System~1 model adopts a deep, narrow configuration
in which the representation is aggressively compressed after the first block and then
refined at a fixed width across four subsequent blocks. 
The System~2 model, by contrast, selects a shallower
but substantially wider architecture.
%
Both models use a SiLU activation function and a
plateau-based \ac{LR} schedule, while differing in their optimizers and batch sizes. The validation accuracy gap between the two models, 85.79\%  for System~1 vs.\ 97.64\% for System~2, indicates
that System~1 presents a more challenging classification task, consistent with the greater spectral overlap between the \textit{eav} and \textit{sbd} signatures observed in the balanced O-band dataset, whereas the filtered and
imbalanced System~2 dataset retains more spectrally
distinct event signatures.

Both final \ac{DNN} models converge without a notable
generalization gap between training and validation sets, with System~1 converging faster (approximately 55~epochs) than System~2 (approximately 115~epochs), consistent with its smaller architecture.
\subsubsection{Best-Found Single-System \ac{VAE} Models}\label{subsubsec: vae sgl hpo rslt}
%
%
Table~\ref{tab:vae_sgl_best} summarizes the
\ac{HPO}-selected configurations for the two
$\text{VAE}_{\text{sgl}}$ models, which reflect the
distinct statistical properties of each system.

\begin{table}[t]
\centering
\caption{\ac{HPO}-selected configuration for the best-found
$\text{VAE}_{\text{sgl}}$ models.}
\label{tab:vae_sgl_best}
\small
\setlength{\tabcolsep}{4pt}
\begin{tabular}{lcc}
\toprule
\textbf{Hyperparameter} & \textbf{System~1} & \textbf{System~2} \\
                        & \textbf{($S_1$, $S_3$)} & \textbf{($S_2$, $S_4$)} \\
\midrule
\multicolumn{3}{l}{\textbf{Architecture}} \\
$N_{vae}$              & 2     & 4 \\
Layer widths           & $512\!\to\!704\!\to\!640\!\to\!32$
                       & $512\!\to\!384\!\to\!320^{\times2}\!\to\!192\!\to\!32$ \\
Latent dim $d_z$       & 32    & 32 \\
Activation $f(\cdot)$  & GELU  & SiLU \\
Batch norm             & Yes   & Yes \\
Dropout $p$            & 0.445 & 0.167 \\
\midrule
\multicolumn{3}{l}{\textbf{Loss Weighting}} \\
\ac{KL} weight $\beta$                    & 0.100 & 0.294 \\
Recon.\ weight $\lambda_{\mathrm{rcst}}$  & 3.886 & 2.530 \\
Classif.\ weight $\lambda_{\mathrm{clf}}$ & 3.248 & 1.414 \\
$\beta$ warm-up (epochs)                  & 43    & 23    \\
\midrule
\multicolumn{3}{l}{\textbf{Optimization}} \\
Optimizer              & AdamW & Adam \\
Learning rate $\eta$   & $2.96\!\times\!10^{-4}$ & $1.62\!\times\!10^{-3}$ \\
Weight decay $\lambda$ & $3.27\!\times\!10^{-6}$ & $9.10\!\times\!10^{-5}$ \\
\ac{LR} schedule       & Plateau ($\times$0.5) & Plateau ($\times$0.5) \\
Batch size             & 256   & 256   \\
\midrule
\multicolumn{3}{l}{\textbf{Result}} \\
Best val.\ accuracy    & 84.75\% & 97.50\% \\
\bottomrule
\end{tabular}
\end{table}
Several contrasts are noteworthy. The lower \ac{DP} for System~2 reflects the need to preserve the scarce
minority-class signal under class imbalance, where
aggressive regularization would suppress the few
\textit{eav} and \textit{sbd} samples. The higher $\beta$ for System~2 indicates that the more complex, imbalanced distribution requires stronger \ac{KL} regularization to maintain a well-structured posterior. The $\beta$ warm-up in both cases confirms that gradual \ac{KL} introduction is
beneficial in both systems, consistent with the known
instability of jointly optimizing reconstruction and
classification objectives from the first epoch. The best validation accuracies of 84.75\% and 97.50\% closely match the corresponding \ac{DNN} baselines shown in Table~\ref{tab:dnn_best}, confirming competitive intra-system discriminability for both
$\text{VAE}_{\text{sgl}}$ models. Both models converge
within approximately 100~epochs without a notable
generalization gap between training and validation sets.
\subsubsection{Best-Found Combined-System \ac{VAE} Model}\label{subsubsec: vae cmb hpo rslt}
%

\begin{table}[t]
\centering
\caption{\ac{HPO}-selected configuration for the best-found $\text{VAE}_{\text{cmb}}$ model.}
\label{tab:vae_cmb_best}
\small
\setlength{\tabcolsep}{4pt}
\begin{tabular}{lc}
\toprule
\textbf{Hyperparameter} & \textbf{Combined ($S_1$--$S_4$)} \\
\midrule
\multicolumn{2}{l}{\textbf{Architecture}} \\
$N_{vae}$            & 3 \\
Layer widths                             & $512\!\to\!128\!\to\!768\!\to\!768\!\to\!32$ \\
Latent dim $d_z$                         & 32 \\
Activation $f(\cdot)$                    & ReLU \\
Batch norm                               & Yes \\
Dropout $p$                              & 0.114 \\
\midrule
\multicolumn{2}{l}{\textbf{Loss Weighting}} \\
\ac{KL} weight $\beta$                   & 0.279 \\
Recon.\ weight $\lambda_{\mathrm{rcst}}$ & 1.161 \\
Classif.\ weight $\lambda_{\mathrm{clf}}$& 2.372 \\
$\beta$ warm-up (epochs)                 & 34 \\
\midrule
\multicolumn{2}{l}{\textbf{Optimization}} \\
Optimizer                                & Adam \\
Learning rate $\eta$                     & $1.87\!\times\!10^{-3}$ \\
Weight decay $\lambda$                   & $1.42\!\times\!10^{-4}$ \\
\ac{LR} schedule                         & cosine \\
Batch size
& 64 \\
\midrule
\multicolumn{2}{l}{\textbf{Result}} \\
Best val.\ accuracy          & 91.79\% \\
\bottomrule
\end{tabular}
\end{table}

In the combined-system \ac{VAE} models, training follows a two-phase design. In Phase~1, the complete model is trained on combined data from both systems, optimizing the combined objective in~\eqref{eq:vae_loss} so that the encoder learns a latent space aligned with the class-discriminative structure consistent with both systems, in accordance with~\eqref{eq:conditional_consistent}. In Phase~2, the encoder is frozen, and a randomly initialized classification head is trained per scenario using only source-domain (single system) labels, with no access to target-domain (cross-system) class information. This design bridges the distribution mismatch in~\eqref{eq:marginal_shift} before any classifier is trained, without requiring target-domain supervision, an advantage structurally unavailable to the \ac{DNN} baseline and previous work in the literature. 
%
%

Table~\ref{tab:vae_cmb_best} summarizes the \ac{HPO}-selected configuration for $\text{VAE}_{\text{cmb}}$, whose single optimization round targets the combined System~1 and System~2 training and validation sets.
The selected 3-layer encoder $512\!\to\!128\!\to\!768\!\to\!768\!\to\!32$ contains 3{,}156{,}099 parameters. The initial compression to 128 units acts as an early domain-agnostic filter before the wider layers learn a shared cross-domain representation. The \ac{KL} weight $\beta\!=\!0.279$, consistent with the System~2 $\text{VAE}_{\text{sgl}}$ (0.294) and higher than the System~1 counterpart (0.100), confirms that combined training demands \ac{KL} regularization comparable to that of the more complex System~2 distribution alone. The classification weight $\lambda_{\mathrm{clf}}\!=\!2.372$ exceeding $\lambda_{\mathrm{rcst}}\!=\!1.161$ reflects deliberate emphasis on class-discriminative pressure. Cosine annealing is selected over the plateau-based schedule of both $\text{VAE}_{\text{sgl}}$ models, providing smoother \ac{LR} decay under the noisier joint-domain loss landscape. The weight decay of $1.42\!\times\!10^{-4}$, one order of magnitude larger than either $\text{VAE}_{\text{sgl}}$ model, reflects the stronger $L_2$ regularization required for the larger parameter count and more diverse training distribution. 
The model converges at epoch~30, achieving a best combined validation accuracy of 91.79\%, with no notable generalization gap between training and validation sets despite the 3.15M-parameter capacity.
\subsection{Experimental Results and Performance Analysis}\label{subsec: performance rslt}

We first compare the overall accuracy of the frameworks across the four scenarios. 
We then analyze its classification behavior in depth through per-scenario confusion matrices and per-class comparisons against the \ac{DNN} baseline.
%
\subsubsection{
Overall Accuracy Comparison over the Four 
Scenarios
}\label{subsubsec: acc cmpr 3 methods}
\begin{figure}[t]
    \centering
    \includegraphics[width=\linewidth]{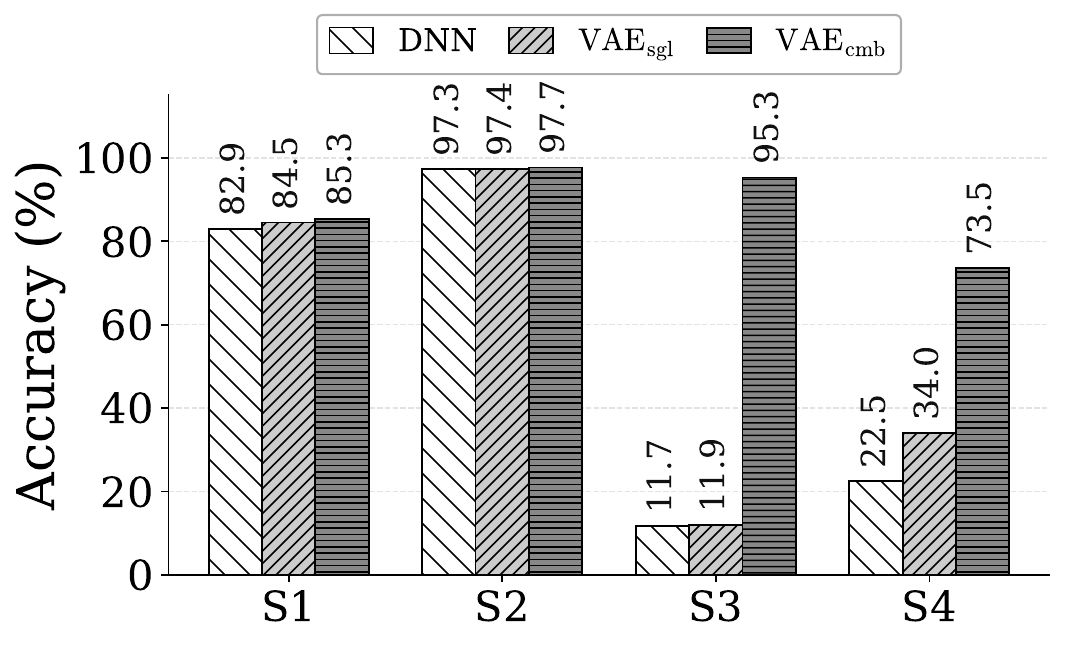}
    \caption{Test-set accuracy comparison across all four evaluation scenarios for the \ac{DNN} baseline, $\text{VAE}_{\text{sgl}}$, and $\text{VAE}_{\text{cmb}}$. 
    }
    \label{fig:acc_cmpr}
\end{figure}

Fig.~\ref{fig:acc_cmpr} compares the test-set accuracy of all three models over all four scenarios. Under intra-system conditions, all three approaches perform comparably.  $\text{VAE}_{\text{cmb}}$ achieves 85.3\% ($S_1$) and 97.7\% ($S_2$) accuracy, marginally surpassing the \ac{DNN} (82.9\% for ($S_1$), 97.3\% for ($S_2$)) and 
$\text{VAE}_{\text{sgl}}$ (84.5\% for ($S_1$), 97.4\% for ($S_2$)). This confirms that combined-domain training does not compromise source-domain discriminability.

The central finding emerges under domain shift. As visible in Fig.~\ref{fig:acc_cmpr}, both the \ac{DNN} and $\text{VAE}_{\text{sgl}}$ collapse well below the 33.3\% random-chance baseline in the cross-system scenarios, reaching only 11.7\% and 11.9\% on $S_3$, and 22.5\% and 34.0\% on $S_4$, respectively. This failure is structural: both models produce domain-specific latent spaces that lack a mechanism to bridge the marginal distribution mismatch in~\eqref{eq:marginal_shift}. $\text{VAE}_{\text{cmb}}$, by contrast, achieves 95.3\% on $S_3$ and 73.5\% on $S_4$, which are absolute gains of 83.4 and 51 percentage points 
over the \ac{DNN} baseline. This demonstrates that combined \ac{VAE} training enables learning a domain-invariant latent representation that supports accurate cross-system classification. 
The remainder of the analysis focuses on the $\text{VAE}_{\text{cmb}}$ performance in detail.

\subsubsection{$\text{VAE}_{\text{cmb}}$ Performance
Analysis}\label{subsubsec: performance vae multiple}

\begin{figure}[t]
    \centering
    \begin{subfigure}[b]{0.48\linewidth}
        \centering
        \includegraphics[width=\linewidth]{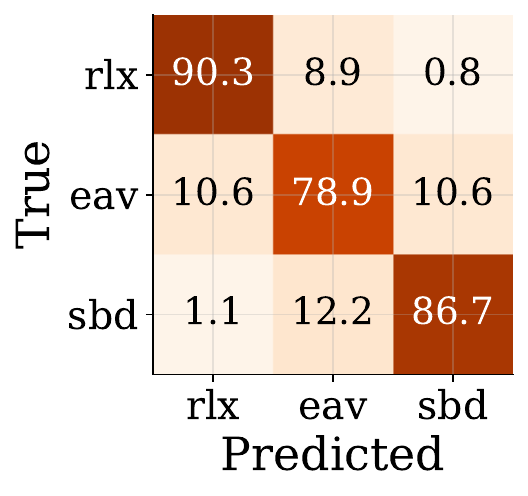}
        \captionsetup{justification=centering}
        \caption{$S_1$: System~1$\!\to\!$System~1}
    \end{subfigure}
    \hfill
    \begin{subfigure}[b]{0.48\linewidth}
        \centering
        \includegraphics[width=\linewidth]{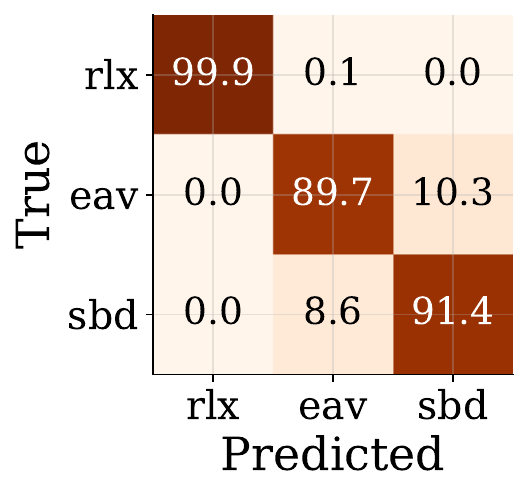}
        \captionsetup{justification=centering}
        \caption{$S_2$: System~2$\!\to\!$System~2}
    \end{subfigure}
    \vspace{4pt}
    \begin{subfigure}[b]{0.48\linewidth}
        \centering
        \includegraphics[width=\linewidth]{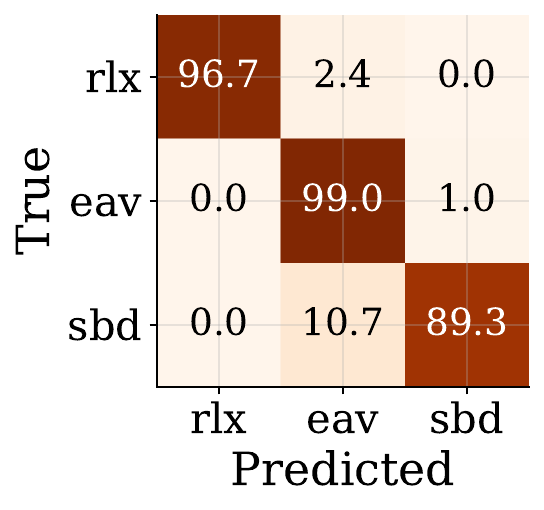}
        \captionsetup{justification=centering}
        \caption{$S_3$: System~1$\!\to\!$System~2}
    \end{subfigure}
    \hfill
    \begin{subfigure}[b]{0.48\linewidth}
        \centering
        \includegraphics[width=\linewidth]{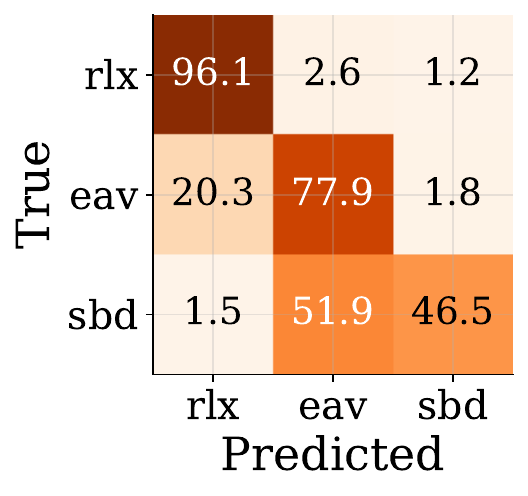}
        \captionsetup{justification=centering}
        \caption{$S_4$: System~2$\!\to\!$System~1}
    \end{subfigure}
    \caption{Confusion matrices of $\text{VAE}_{cmb}$ for all four evaluation scenarios. Values are per-class accuracy (\%). 
    }
    \label{fig:cms}
\end{figure}

\paragraph{Scenario-Level Analysis via Confusion Matrices.}
Fig.~\ref{fig:cms} reports the confusion matrices of $\text{VAE}_{\text{cmb}}$ for all four scenarios. In the intra-system scenarios, accuracy for $S_1$ is high across all classes, with the modest \textit{eav} prediction rate drop.
In the intra-system scenario $S_2$, the model achieves near-perfect performance in detecting the normal state, but we still observe misclassifications between \textit{eav} and \textit{sbd}.

For the highly-challenging cross-system scenario $S_3$, $\text{VAE}_{cmb}$ classifies \textit{rlx} and \textit{eav} with near-perfect accuracy (97.6\% and 99.0\%). \textit{sbd} is correctly detected in 89.3\% of cases at a residual confusion with \textit{eav} (10.7\%), likely reflecting the inherent spectral similarity between gradual bending and eavesdropping. 
Nonetheless, this represents a complete recovery from the baseline's low performance. 
In $S_4$, \textit{rlx} is classified with high accuracy of 96.1\% and \textit{eav} with 77.9\%. Here, \textit{sbd} at only 46.5\% accuracy remains the most challenging class, with 51.9\% of samples misclassified as \textit{eav}. 

\begin{figure}[t]
    \centering
    \includegraphics[width=\linewidth]{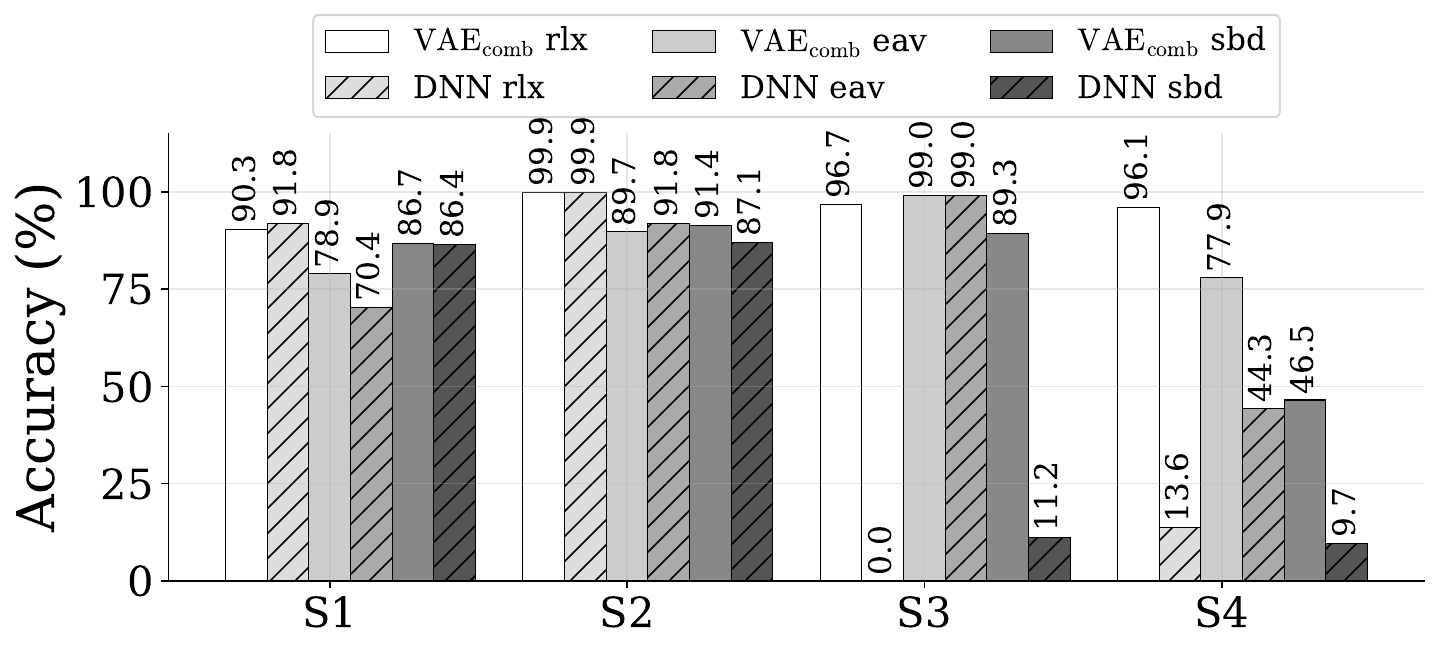}
    \caption{Per-class accuracy of $\text{VAE}_{\text{cmb}}$ vs.\ the \ac{DNN} baseline across all four scenarios. 
    }
    \label{fig:perclass_vae}
\end{figure}

\paragraph{Per-Class Comparison with the \ac{DNN} Baseline.}
Fig.~\ref{fig:perclass_vae} compares the per-class accuracy of $\text{VAE}_{\text{cmb}}$ and the \ac{DNN} baseline across all four scenarios. In the intra-system settings ($S_1$, $S_2$), both models achieve comparable per-class rates, with $\text{VAE}_{cmb}$ providing slight improvements, particularly for the minority classes \textit{eav} and \textit{sbd}, suggesting that the regularized latent space marginally improves generalization even within a single domain. However, the per-class breakdown under cross-system conditions ($S_3$, $S_4$) highlights the fundamental research gap addressed by this work: the \ac{DNN} baseline achieves near-zero accuracy on \textit{rlx} and \textit{sbd} in $S_3$, and severely degraded rates across all classes in $S_4$. 
$\text{VAE}_{cmb}$ recovers all three classes to operationally meaningful levels of accuracy in both cases, with the \textit{sbd} class in $S_4$  remaining the primary challenge for future improvement. 
%
This result establishes $\text{VAE}_{cmb}$ as an effective solution to the cross-system generalization problem for \ac{SOP}-based monitoring of physical-layer security breaches.

%

The primary remaining challenge is the \textit{sbd}--\textit{eav} confusion in the System~2$\!\to\!$System~1 direction, reflecting the greater representational complexity of the live metro ring relative to the dark fiber testbed. 
\section{Conclusion}\label{sec: conclusion}
This paper introduced a \ac{VAE}-based \ac{DA} framework for cross-system generalization in \ac{ML}-driven \ac{SOP}-based fiber monitoring, addressing the failure of existing supervised approaches to transfer across optical systems with different spectral bands, fiber links, and network topologies.
Three models were evaluated over four intra- and cross-system scenarios using real-world \ac{SOP} data from two fundamentally different optical links and topologies. While all models achieved comparable intra-system accuracy, both the supervised \ac{DNN} baseline and $\text{VAE}_{\text{sgl}}$ collapsed to below random-chance accuracy under cross-system transfer. By contrast, the proposed $\text{VAE}_{\text{cmb}}$, trained on combined data from both systems through a two-phase design, achieved meaningful accuracy in both transfer directions.

\section*{Funding and Acknowledgments}
This work was supported by the European Union’s Horizon Europe research and innovation programme (Grant No. 10113933) through the ICON project and the Digital Europe Programme (Grant No. 101127973) through the 5G-TACTIC project,
as well as by the CELTIC-NEXT SUSTAINET-Advance project, Taighde Éireann – Research Ireland under Grant No. 18/RI/5721 (OpenIreland Research Infrastructure), 13/RC/2106\_P2 (ADAPT centre) and the Swedish Research Council (Grant No. 2023–05249).




\bibliography{ref}

\begin{thebibliography}{10}
\newcommand{\enquote}[1]{``#1''}

\bibitem{rafique_jlt_2018}
D.~Rafique, T.~Szyrkowiec, H.~Grie{\ss}er, A.~Autenrieth, and J.-P. Elbers, \enquote{Cognitive assurance architecture for optical network fault management,} {\protect\JournalTitle{Journal of Lightwave Technology}} \textbf{36}, 1443--1450 (2018).

\bibitem{allwood_2016_Sensors_Journal}
G.~Allwood, G.~Wild, and S.~Hinckley, \enquote{Optical fiber sensors in physical intrusion detection systems: A review,} {\protect\JournalTitle{IEEE Sensors Journal}} \textbf{16}, 5497--5509 (2016).

\bibitem{pellegrini_jocn_2025}
S.~Pellegrini, L.~Minelli, L.~Andrenacci, G.~Rizzelli, D.~Pilori, G.~Bosco, L.~Della~Chiesa, C.~Crognale, S.~Piciaccia, and R.~Gaudino, \enquote{Overview on the state of polarization sensing: Application scenarios and anomaly detection algorithms,} {\protect\JournalTitle{Journal of Optical Communications and Networking}} \textbf{17}, A196--A209 (2025).

\bibitem{lu_2019_APR}
P.~Lu, N.~Lalam, M.~Badar \emph{et~al.}, \enquote{Distributed optical fiber sensing: Review and perspective,} {\protect\JournalTitle{Applied Physics Reviews}} \textbf{6} (2019).

\bibitem{carver_2024_nature}
C.~J. Carver and X.~Zhou, \enquote{Polarization sensing of network health and seismic activity over a live terrestrial fiber-optic cable,} {\protect\JournalTitle{Communications Engineering}} \textbf{3}, 91 (2024).

\bibitem{ls_ofc_2024}
L.~Sadighi, S.~Karlsson, C.~Natalino, and M.~Furdek, \enquote{Machine learning-based polarization signature analysis for detection and categorization of eavesdropping and harmful events,} in \emph{Optical Fiber Communications Conference and Exhibition (OFC),}  (2024), p. M1H.1.

\bibitem{ls_ecoc_2025}
L.~Sadighi, C.~Natalino, S.~Karlsson, M.~Ruffini, E.~Kenny, L.~Wosinska, and M.~Furdek, \enquote{Generalizability of{ ML}-based classification of state of polarization signatures across different bands and links,} in \emph{2025 European Conference on Optical Communications (ECOC),}  (2025), pp. 1--4.

\bibitem{pan_tkde_2010}
S.~J. Pan and Q.~Yang, \enquote{A survey on transfer learning,} {\protect\JournalTitle{IEEE Transactions on Knowledge and Data Engineering}} \textbf{22}, 1345--1359 (2010).

\bibitem{kingma_iclr_2014}
D.~P. Kingma and M.~Welling, \enquote{Auto-encoding variational {B}ayes,} in \emph{2nd International Conference on Learning Representations ({ICLR}),}  (Banff, AB, Canada, 2014).

\bibitem{rode_oft_2025}
A.~Rode, M.~Farsi, V.~Lauinger, M.~Karlsson, E.~Agrell, L.~Schmalen, and C.~H{\"a}ger, \enquote{Machine learning opportunities for integrated polarization sensing and communication in optical fibers,} {\protect\JournalTitle{Optical Fiber Technology}} \textbf{89}, 103924 (2025).

\bibitem{ls_icton_2024}
L.~Sadighi, S.~Karlsson, L.~Wosinska, and M.~Furdek, \enquote{Machine learning analysis of polarization signatures for distinguishing harmful from non-harmful fiber events,} in \emph{International Conference on Transparent Optical Networks (ICTON),}  (2024).

\bibitem{ls_ecoc_2024}
L.~Sadighi, S.~Karlsson, C.~Natalino, L.~Wosinska, M.~Ruffini, and M.~Furdek, \enquote{Detection and classification of eavesdropping and mechanical vibrations in fiber optical networks by analyzing polarization signatures over a noisy environment,} in \emph{European Conference on Optical Communication (ECOC),}  (2024), pp. 527--530.

\bibitem{ls_JLT_2025}
L.~Sadighi, S.~Karlsson, C.~Natalino, L.~Wosinska, M.~Ruffini, and M.~Furdek, \enquote{Deep learning for detection of harmful events in real-world, noisy optical fiber deployments,} {\protect\JournalTitle{J. Lightwave Technol.}} \textbf{43}, 6092--6101 (2025).

\bibitem{tomasov_olt_2023}
A.~Tomasov, P.~Dejdar, P.~Munster, T.~Horvath, P.~Barcik, and F.~Da~Ros, \enquote{Enhancing fiber security using a simple state of polarization analyzer and machine learning,} {\protect\JournalTitle{Optics \& Laser Technology}} \textbf{167}, 109668 (2023).

\bibitem{abdelli_jlt_2025}
K.~Abdelli, M.~Lonardi, J.~Gripp, D.~Correa, S.~Olsson, F.~Boitier, and P.~Layec, \enquote{Vision transformers for anomaly classification and localization in optical networks using {SOP} spectrograms,} {\protect\JournalTitle{Journal of Lightwave Technology}}  (2025).

\bibitem{karlsson_ofc_2023}
S.~Karlsson, M.~Andersson, R.~Lin, L.~Wosinska, and P.~Monti, \enquote{Detection of abnormal activities on a {SM} or {MM} fiber,} in \emph{Optical Fiber Communication Conference (OFC),}  (Optica Publishing Group, San Diego, CA, USA, 2023), p. M3Z.6.

\bibitem{qin_ucom_2024}
W.~Qin, Q.~Zhang, W.~Hou, X.~Zhang, and X.~Gong, \enquote{Convolutional neural networks for fiber-bending eavesdropping attacks detection in coherent optical communication systems,} in \emph{International Conference on Ubiquitous Communication (Ucom),}  (Xi'an, China, 2024).

\bibitem{ls_tnsm_2025}
L.~Sadighi, S.~Karlsson, C.~Natalino, and M.~Furdek, \enquote{Ml-based state of polarization analysis to detect emerging threats to optical fiber security,} {\protect\JournalTitle{IEEE Transactions on Network and Service Management}} \textbf{23}, 432--442 (2026).

\bibitem{song_jocn_2024}
H.~Song, R.~Lin, L.~Wosinska, P.~Monti, M.~Zhang, Y.~Liang, Y.~Li, and J.~Zhang, \enquote{Cluster-based unsupervised method for eavesdropping detection and localization in {WDM} systems,} {\protect\JournalTitle{Journal of Optical Communications and Networking}} \textbf{16}, F52--F61 (2024).

\bibitem{minelli_ecoc_2023}
L.~Minelli, S.~Pellegrini, L.~Andrenacci, D.~Pilori, G.~Bosco, L.~Della~Chiesa, A.~Tanzi, C.~Crognale, and R.~Gaudino, \enquote{{SOP}-based {DSP} blind anomaly detection for sensing on deployed metropolitan fibers,} in \emph{European Conference on Optical Communications (ECOC),}  (Glasgow, UK, 2023), pp. 519--522.

\bibitem{rottondi_jocn_2021}
C.~Rottondi, R.~di~Marino, M.~Nava, A.~Giusti, and A.~Bianco, \enquote{On the benefits of domain adaptation techniques for quality of transmission estimation in optical networks,} {\protect\JournalTitle{Journal of Optical Communications and Networking}} \textbf{13}, A34--A43 (2021).

\bibitem{musumeci_jocn_2022}
F.~Musumeci, V.~Garbhapu~Venkata, Y.~Hirota, Y.~Awaji, S.~Xu, M.~Shiraiwa, B.~Mukherjee, and M.~Tornatore, \enquote{Domain adaptation and transfer learning for failure detection and failure-cause identification in optical networks across different lightpaths [invited],} {\protect\JournalTitle{Journal of Optical Communications and Networking}} \textbf{14}, A91--A100 (2022).

\bibitem{8386174}
W.~Mo, Y.-K. Huang, S.~Zhang, E.~Ip, D.~C. Kilper, Y.~Aono, and T.~Tajima, \enquote{Ann-based transfer learning for qot prediction in real-time mixed line-rate systems,} in \emph{2018 Optical Fiber Communications Conference and Exposition (OFC),}  (2018), pp. 1--3.

\bibitem{yu_jocn_2019}
J.~Yu, W.~Mo, Y.-K. Huang, E.~Ip, and D.~C. Kilper, \enquote{Model transfer of {QoT} prediction in optical networks based on artificial neural networks,} {\protect\JournalTitle{Journal of Optical Communications and Networking}} \textbf{11}, C48--C57 (2019).

\bibitem{cai_jocn_2024}
Z.~Cai, Q.~Wang, Y.~Deng, P.~Zhang, G.~Zhou, Y.~Li, and F.~N. Khan, \enquote{Domain adversarial adaptation framework for few-shot {QoT} estimation in optical networks,} {\protect\JournalTitle{Journal of Optical Communications and Networking}} \textbf{16}, 1133--1144 (2024).

\bibitem{abdelli_jocn_2024}
K.~Abdelli, M.~Lonardi, J.~Gripp, S.~Olsson, F.~Boitier, and P.~Layec, \enquote{Risky event classification leveraging transfer learning for very limited datasets in optical networks,} {\protect\JournalTitle{Journal of Optical Communications and Networking}} \textbf{16}, C51--C68 (2024).

\bibitem{ilse_midl_2020}
M.~Ilse, J.~M. Tomczak, C.~Louizos, and M.~Welling, \enquote{{DIVA}: Domain invariant variational autoencoders,} in \emph{Proceedings of the Third Conference on Medical Imaging with Deep Learning,}  vol. 121 of \emph{Proceedings of Machine Learning Research} (PMLR, 2020), pp. 322--348.

\bibitem{louizos_iclr_2016}
C.~Louizos, K.~Swersky, Y.~Li, M.~Welling, and R.~S. Zemel, \enquote{The variational fair autoencoder,} in \emph{4th International Conference on Learning Representations ({ICLR}),}  (2016).

\bibitem{hsu_asru_2017}
W.-N. Hsu, Y.~Zhang, and J.~Glass, \enquote{Unsupervised domain adaptation for robust speech recognition via variational autoencoder-based data augmentation,} in \emph{IEEE Automatic Speech Recognition and Understanding Workshop (ASRU),}  (IEEE, 2017), pp. 16--23.

\bibitem{OIR_2024}
{CONNECT Centre for Future Networks and Communications}, \enquote{{Open Ireland Testbed},}  (2020). Available \href{https://connectcentre.ie/news/launch-of-open-ireland-e2-million-open-networking-testbed/}{here}.

\bibitem{HEAnet}
{ASIERA (formerly HEAnet)}, \enquote{Ireland’s {N}ational {E}ducation and {R}esearch {N}etwork,}  (2025). Available \href{https://www.heanet.ie/}{here}.

\bibitem{Karlsson_patent}
S.~Karlsson, \enquote{A method for detecting external influence on an optical cable,}  (1994). Filed Mar. 9, 1989; granted Oct. 19, 1994.

\bibitem{podder_2014_hamming}
P.~Podder, T.~Z. Khan, M.~H. Khan, and M.~M. Rahman, \enquote{Comparative performance analysis of hamming, hanning and blackman window,} {\protect\JournalTitle{International Journal of Computer Applications}} \textbf{96}, 1--7 (2014).

\bibitem{van_2008_jml}
L.~Van~der Maaten and G.~Hinton, \enquote{Visualizing data using t-sne.} {\protect\JournalTitle{Journal of machine learning research}} \textbf{9} (2008).

\bibitem{BN_2015_ICML}
S.~Ioffe and C.~Szegedy, \enquote{Batch normalization: Accelerating deep network training by reducing internal covariate shift,} in \emph{Proceedings of the 32nd International Conference on Machine Learning (ICML),}  (PMLR, 2015), pp. 448--456.

\bibitem{DP_2014_JMLR}
N.~Srivastava, G.~Hinton, A.~Krizhevsky, I.~Sutskever, and R.~Salakhutdinov, \enquote{Dropout: A simple way to prevent neural networks from overfitting,} {\protect\JournalTitle{Journal of Machine Learning Research (JMLR)}} \textbf{15}, 1929--1958 (2014).

\bibitem{lin_2017_focal}
T.-Y. Lin, P.~Goyal, R.~Girshick, K.~He, and P.~Doll{\'a}r, \enquote{Focal loss for dense object detection,} in \emph{Proceedings of the IEEE International Conference on Computer Vision (ICCV),}  (IEEE, 2017), pp. 2980--2988.

\bibitem{optuna_2019_ACM}
T.~Akiba, S.~Sano, T.~Yanase, T.~Ohta, and M.~Koyama, \enquote{{Optuna}: A next-generation hyperparameter optimization framework,} in \emph{Proceedings of the 25th ACM SIGKDD International Conference on Knowledge Discovery and Data Mining (KDD),}  (ACM, 2019), pp. 2623--2631.

\bibitem{Pytorch_2019_NeurIPS}
A.~Paszke, S.~Gross, F.~Massa, A.~Lerer, J.~Bradbury, G.~Chanan, T.~Killeen, Z.~Lin, N.~Gimelshein, L.~Antiga \emph{et~al.}, \enquote{{PyTorch}: An imperative style, high-performance deep learning library,} in \emph{Advances in Neural Information Processing Systems (NeurIPS),}  vol.~32 (Curran Associates, Inc., 2019).

\end{thebibliography}

\end{document}